\documentclass[letterpaper]{article} 
\usepackage{aaai2026}  
\usepackage{times}  
\usepackage{helvet}  
\usepackage{courier}  
\usepackage[hyphens]{url}  
\usepackage{graphicx} 
\usepackage{natbib}  
\usepackage{caption} 
\usepackage{algorithm}
\usepackage{algorithmic}

\usepackage{booktabs}       
\usepackage{amsfonts}       
\usepackage{nicefrac}       
\usepackage{microtype}      

\usepackage{lipsum}		
\usepackage{color}
\usepackage{arydshln}
\usepackage{amsmath}
\graphicspath{ {./images/} }
\usepackage{multirow}       
\usepackage{listings}       
\usepackage{tcolorbox}      
\usepackage[table]{xcolor}  
\usepackage{makecell} 
\usepackage{multicol}

\usepackage{newfloat}
\usepackage{listings}
\DeclareCaptionStyle{ruled}{labelfont=normalfont,labelsep=colon,strut=off} 
\floatstyle{ruled}
\newfloat{listing}{tb}{lst}{}
\floatname{listing}{Listing}

\definecolor{LightGray}{gray}{0.99}
\definecolor{grey1}{RGB}{247,202,173}
\definecolor{grey2}{RGB}{255,241,204}
\definecolor{grey3}{RGB}{225,240,217}
\definecolor{grey4}{RGB}{218,227,243}
\definecolor{grey5}{RGB}{244,247,251}

\newcommand{\prompt}[1]{\rowcolor{grey3!30} #1  \\}
\newcommand{\functionfirst}[1]{\rowcolor{grey2!10} #1  \\}
\newcommand{\functionsecond}[1]{\rowcolor{grey2!10}  #1  \\}
\newcommand{\breadthquestion}[1]{\rowcolor{grey4!70}  #1  \\}
\newcommand{\breadthanswer}[1]{\rowcolor{grey4!30}  #1  \\}
\newcommand{\depthreference}[1]{\rowcolor{grey2!10}  #1  \\}

\title{Thinker: Training LLMs in Hierarchical Thinking for Deep Search via Multi-Turn Interaction}
\author{
    Jun Xu\textsuperscript{\rm 1},
    Xinkai Du\textsuperscript{\rm 1},
    Yu Ao\textsuperscript{\rm 1},
    Peilong Zhao\textsuperscript{\rm 1},
    Yang Li\textsuperscript{\rm 1},
    Ling Zhong\textsuperscript{\rm 1},
    Lin Yuan\textsuperscript{\rm 1},
    Zhongpu Bo\textsuperscript{\rm 1},
    Xiaorui Wang\textsuperscript{\rm 1},
    Mengshu Sun\textsuperscript{\rm 1},
    Zhengke Gui\textsuperscript{\rm 1},
    Dalong Zhang\textsuperscript{\rm 1},
    Zhaoyang Wang\textsuperscript{\rm 1},
    Qiwei Wang\textsuperscript{\rm 1},
    Yangyang Hou\textsuperscript{\rm 1},
    Zhiying Yin\textsuperscript{\rm 1},
    Haofen Wang\textsuperscript{\rm 2},
    Huajun Chen\textsuperscript{\rm 3},
    Lei Liang\textsuperscript{\rm 1}$^\ast$,
    Jun Zhou\textsuperscript{\rm 1}\thanks{Corresponding author.}
}
\affiliations{
    \textsuperscript{\rm 1}Ant Group, Hangzhou, China\\
    \textsuperscript{\rm 2}Tongji University, Shanghai, China\\
    \textsuperscript{\rm 3}Zhejiang University, Hangzhou, China\\
    \{xujun.xj, duxinkai.dxk, mengshu.sms, leywar.liang, jun.zhoujun\}@antgroup.com
}

\usepackage{bibentry}

\begin{document}

\maketitle

\begin{abstract}
Efficient retrieval of external knowledge bases and web pages is crucial for enhancing the reasoning abilities of LLMs.
Previous works on training LLMs to leverage external retrievers for solving complex problems have predominantly employed end-to-end reinforcement learning. However, these approaches neglect supervision over the reasoning process, making it difficult to guarantee logical coherence and rigor.
To address these limitations, we propose Thinker, a hierarchical thinking model for deep search through multi-turn interaction, making the reasoning process supervisable and verifiable.
It decomposes complex problems into independently solvable sub-problems, each dually represented in both natural language and an equivalent logical function to support knowledge base and web searches. 
Concurrently, dependencies between sub-problems are passed as parameters via these logical functions, enhancing the logical coherence of the problem-solving process. 
To avoid unnecessary external searches, we perform knowledge boundary determination to check if a sub-problem is within the LLM's intrinsic knowledge, allowing it to answer directly.
Experimental results indicate that with as few as several hundred training samples, the performance of Thinker is competitive with established baselines. 
Furthermore, when scaled to the full training set, Thinker significantly outperforms these methods across various datasets and model sizes. The source code
is available at https://github.com/OpenSPG/KAG-Thinker.
\end{abstract}

\section{Introduction}
Recently, large language models (LLMs) like GPT-4 have demonstrated remarkable capabilities dealing with simple tasks~\cite{openai2024gpt4technicalreport,qwen2025qwen25technicalreport}. However, they often fall short when handling complex tasks that require in-depth investigation, information synthesis, and multi-step reasoning. They not only face challenges in planning solution pathways and integrating information from diverse sources but are also prone to hallucinations (confidently fabricating information when their knowledge is insufficient). 
\begin{figure}[htbp]
    \centering
    \includegraphics[width=\linewidth]{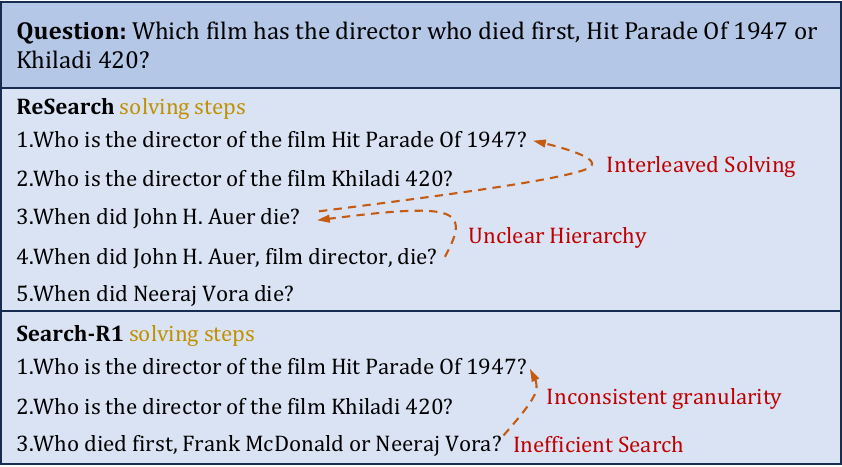}
    \caption{Typical problems with deep search methods based on reinforcement learning training.}
    \label{fig:introduction}
    \vspace{-0.15cm}
\end{figure}
To address these limitations, a variety of methods~\cite{wang2025stepsearchignitingllmssearch,chen2025researchlearningreasonsearch,sun2025zerosearchincentivizesearchcapability,jin2025searchr1trainingllmsreason,zhang2025kagthinkerinteractivethinkingdeep,10.1145/3701716.3715240} have been proposed that focus on emulating a deep search process. Most training-based deep search models are primarily built on reinforcement learning. However, reinforcement learning struggles to constrain the problem-solving process, especially for complex or hybrid tasks. As illustrated in Figure~\ref{fig:introduction}, reinforcement learning based deep search methods often suffer from typical problems such as interleaved solving, unclear hierarchy, inconsistent granularity, and inefficient search. These methods often resemble an unstructured, freestyle approach. The reasoning process lacks a systematic framework, is not logically rigorous, and yields inconsistent results. In stark contrast, human experts solve complex problems using a structured thinking process: they first decompose a large problem into smaller, independent sub-problems and then tackle them one by one~\cite{DBLP:journals/corr/abs-2505-02665,DBLP:conf/emnlp/HuaZ22}. 
Inspired by this observation, we propose a hierarchical thinking framework to address the limitations of existing deep search methods, improving the logic and rigor of the problem-solving process.

\noindent\textbf{Breadth Decomposition and Depth Solving}.
Direct retrieval often proves insufficient for complex multi-hop problems, as it fails to provide immediate answers. Consider, for example, the question:\textit{``Which film has the director who died first, Hit Parade Of 1947 or Khiladi 420?''}. To address such issues, we have devised a model based on breadth decomposition and depth solving. At the breadth-level, the decomposition of complex problems into sub-problems is generally straightforward and typically does not require external knowledge. For instance, the aforementioned example can be decomposed into five constituent sub-problems: \textit{``Who is the director of Hit Parade Of 1947?'', ``When did \#1 die?'', ``Who is the director of Khiladi 420?'', ``When did \#3 die?''}, and \textit{``Which film was directed by the director who died first according to \#2 and \#4?''}. At the depth-level, the resolution of each individual sub-problem may require anywhere from $0$ to $M$ (maximum number of searches) retrieval operations, contingent upon its inherent complexity.
The in-depth approach provides a mechanism for augmenting sub-problems with requisite external knowledge, a process crucial for achieving their solvability.

\noindent\textbf{Knowledge Boundary Determination}. Most prior approaches~\cite{asai2023selfraglearningretrievegenerate,DBLP:conf/nips/GutierrezS0Y024} initiate retrieval for every problem or sub-problem, even when the required information is already encoded in the LLM's parametric knowledge. To prevent unnecessary and potentially noisy retrieval, we introduce a knowledge boundary determination module. Prior to knowledge boundary determination, the LLM first generates answers to the sub-problems. To ensure the precision of this knowledge boundary determination, we employ two strategies: (1) prompt-based confidence assessment and (2) likelihood-based confidence assessment. Only when both conditions are satisfied is a sub-problem deemed solvable without external retrieval, and the answer generated by the LLM is directly adopted.

\noindent\textbf{Dual Representation based Reasoning}. Prior deep search approaches~\cite{wang2025stepsearchignitingllmssearch,chen2025researchlearningreasonsearch,sun2025zerosearchincentivizesearchcapability,jin2025searchr1trainingllmsreason,DBLP:conf/nips/LewisPPPKGKLYR020,Search-o1}, when handling both problems and sub-problems, are restricted to plain text representations, thereby preventing them from effectively utilizing high-quality structured knowledge bases. To address this issue, we introduce four logical forms tailored to frequently encountered problems (see Appendix~\ref{appendix:logical_form}). Each logical form comprises two components: a natural language segment (Step) and a logical function expression segment (Action). For generic retrievers like E5 and BGE-M3, we can directly employ the content within the logical form's Step. Conversely, for structured knowledge retrievers, the Action portion of the logical form can be utilized. Crucially, the logical form facilitates the propagation of dependencies among interdependent sub-problems. For example, consider \textit{``Step1: Who is the director of Hit Parade Of 1947? Action1: Retrieval(s=s1:film[`Hit Parade Of 1947'], p=p1:director, o=o1:director)''} followed by \textit{``Step2: When did \#1 die? Action2: Retrieval(s=o1, p=p2:deathtime, o=o2:deathtime)''}. The variables $\#1$ and $o1$ enable seamless transfer of text and expressions from the outcome of the first planning step. In this manner, the logical form enhances the logic, rigor, and stability of the LLM's reasoning.

In summary, the main contribution of this work is as follows:
\begin{itemize}
\item We introduce a hierarchical deep search method that enhances logical rigor in complex problem-solving through multi-turn interaction. The method innovatively employs a dual representation of natural language and logical functions for effective knowledge base and web retrieval during breadth decomposition and depth solving, while a knowledge boundary detection module minimizes unnecessary external searches to boost overall performance.
\item Our model significantly outperforms established baselines across various datasets and model sizes. Furthermore, our approach demonstrates remarkable sample efficiency, achieving performance competitive with established baselines while using only a few hundred training samples.
\end{itemize}

\section{Related Work}
LLMs often lack domain-specific knowledge and are prone to hallucinations. To mitigate these limitations, external knowledge bases and search engines are widely integrated to supply external information. Several retrieval-augmented generation (RAG) methods have been proposed~\cite{DBLP:conf/naacl/JeongBCHP24,DBLP:conf/emnlp/IslamRHHJP24,DBLP:conf/coling/TangGLDLX25,DBLP:conf/icml/BorgeaudMHCRM0L22,DBLP:conf/cikm/DongLWZXX23,DBLP:conf/acl/LuoETPG0MDSLZL24,Li2025FromS1,DBLP:journals/corr/abs-2502-06772,DBLP:journals/corr/abs-2402-11035}. However, these methods are reliant on a monolithic, single-pass retrieval step, which often yields responses based on incomplete or superficial evidence. To mitigate these shortcomings, deep search methods have been developed. These methods enhance RAG by operationalizing an iterative search-read-reason-refine cycle. This process persists until a predefined termination criterion is met, thereby optimizing for more comprehensive and accurate outcomes. Existing deep search strategies can be broadly categorized into two paradigms~\cite{DBLP:conf/naacl/JeongBCHP24,DBLP:conf/emnlp/IslamRHHJP24,DBLP:conf/coling/TangGLDLX25,wang2025stepsearchignitingllmssearch,chen2025researchlearningreasonsearch,sun2025zerosearchincentivizesearchcapability,DBLP:journals/corr/abs-2411-18478}: (1) API-driven systems that leverage task decomposition and planning, often employing multi-agent architectures to orchestrate retrieval decisions; and (2) Interaction-trained models, where LLMs are fine-tuned, typically via reinforcement learning, for more seamless and effective retriever interaction. Despite these innovations, a paramount challenge lies in the opacity of their intermediate reasoning steps, which impedes verifiability and can compromise the logical integrity of the solution path. 

\begin{figure*}[ht]
\centering 
\includegraphics[width=\textwidth]{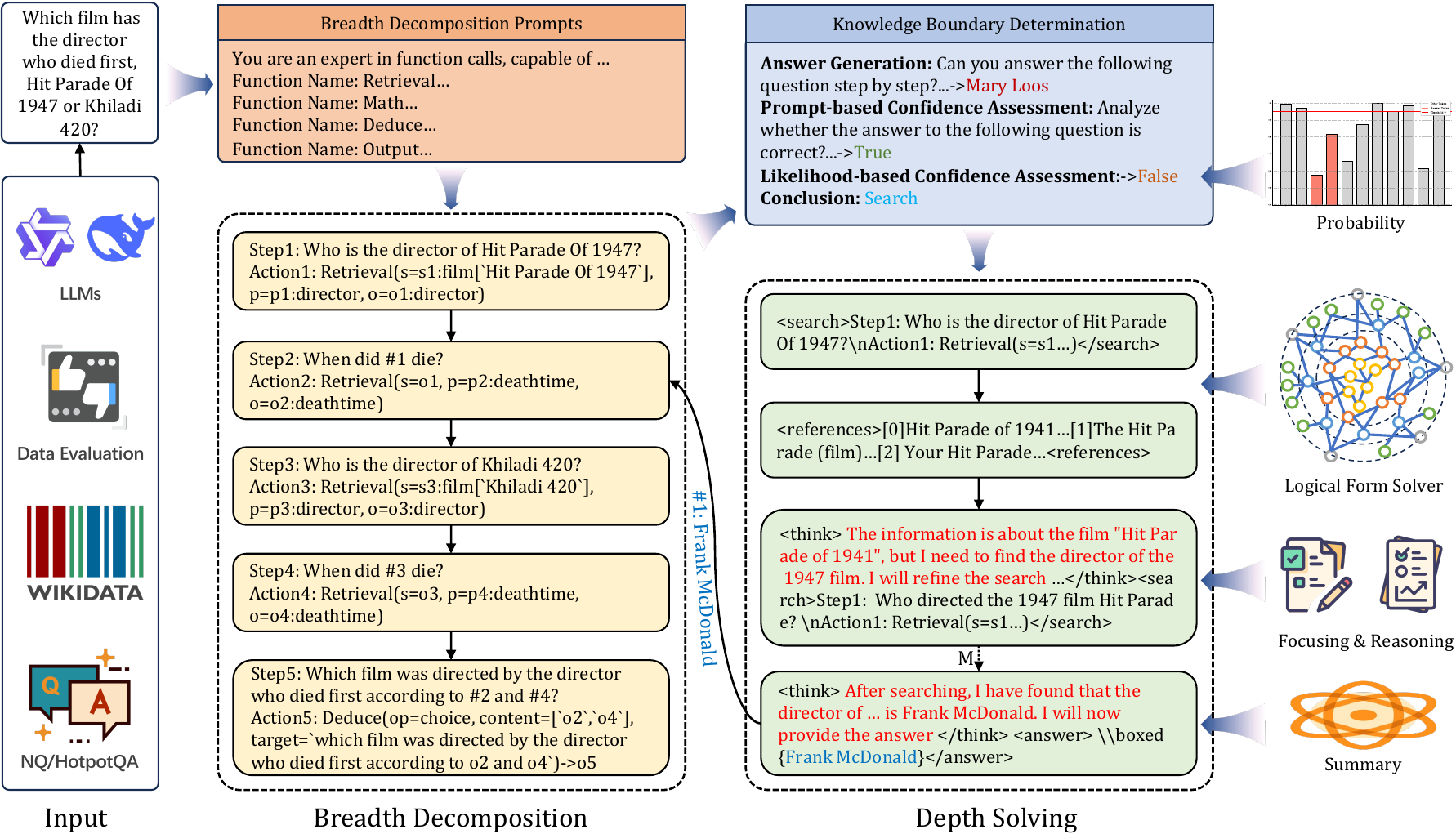} 
\caption{Overview of the Thinker model that uses hierarchical thinking through a multi-turn reasoning process. During problem breadth decomposition, all sub-problems are obtained in a single decomposition pass, where each sub-problem is an atomic problem that can be solved independently. Herein, the terms \textbf{Step} and \textbf{Action} maintain semantic consistency, both denoting such a sub-problem. Within problem breadth decomposition, Step employs $\#n$ for answer propagation of the $n$-th sub-problem, while Action binds variables in logical function (e.g., $o_n$, $s_n$) for variable transmission. By determining the knowledge boundary of the sub-problem, it is decided whether to utilize the base model's answer or to generate a deep retrieval. During depth solving of sub-problems, the system sequentially executes retrieval, focusing, and reasoning in iterative processes until either the sub-problem answer is obtained or the maximum solving attempt threshold is activated.} 
\label{Fig.KAGThinker_framework} 
\end{figure*}
\section{Approach}
We propose a deep search method based on hierarchical thinking and multi-turn interaction, as shown in Figure~\ref{Fig.KAGThinker_framework}. 

\subsection{Breadth Decomposition and Depth Solving}
Complex multi-hop problems usually need to be broken down to be solved. 
Our method decomposes problems into two parts: breadth decomposition, which ensures that the main problem and sub-problems remain logical and precise, and in-depth solving, which ensures that sub-problems are provided with sufficient knowledge to be solved.

\noindent\textbf{Breadth Decomposition}
We decompose complex problems breadth-wise into $n$ atomic granularity sub-problems, with our decomposition instruction template detailed in Appendix~\ref{appendix.breadth_decomposition}. We define four logical form functions (Retrieval, Math, Deduce, and Output), each dedicated to handling specific tasks: retrieval-focused problems, mathematical computation and causal reasoning tasks, and results aggregation functions. For more details, please see Appendix~\ref{appendix:logical_form}. As shown in Figure~\ref{Fig.KAGThinker_framework}, our breadth decomposition divides the original question into five logical forms of atomic granularity, each independently solvable. The dependency variables are propagated between logical forms by function variable $\#n$, $o_n$, and $s_n$. This approach ensures logically coherent question-solving while maintaining consistent sub-question granularity. To ensure compatibility with both natural language and logical form retrievers during sub-question solving, each sub-question adopts dual representations, \textbf{Step} and \textbf{Action}, that maintain semantic equivalence.

\noindent\textbf{Depth Solving}
Constrained by sub-problem representations and retriever capabilities, many sub-problems defy resolution through single-turn retrievals. We have engineered a depth-oriented resolution strategy for the \textit{Retrieval} sub-problem, as detailed in the right part of Figure~\ref{Fig.KAGThinker_framework}. During depth solving, we prompt the LLM to iteratively search across multi-level and multi-dimensional contexts. The iterative process continues until one of the following conditions is met: (1) the maximum number of turns is reached, or (2) the model generates a conclusive response enclosed between the designated answer tokens $<$answer$>$ and $<$/answer$>$. Before conducting depth solving of the \textit{Retrieval} sub-problem, we first perform a knowledge boundary determination on the sub-problem to assess whether the LLM can directly answer it. If the LLM can confidently answer the current sub-problem, there is no need for depth solving, and the answer can be directly obtained; otherwise, a depth solving process will be initiated. After retrieving relevant content, we conduct focusing and reasoning analysis, allowing the model to determine contextually whether to initiate the next action. If a subsequent action is required, a new logical form is generated; otherwise, the sub-question answer is produced. This process iterates interactively until either the maximum number of turns is reached or the sub-question answer is obtained.

\subsection{Knowledge Boundary Determination} 
Exhaustive retrieval for every sub-problem introduces significant computational overhead and noise, causing more hallucinations. To mitigate these challenges, we introduce the \textbf{Generate First, Then Assess} strategy. This approach mandates that the model first attempts to formulate a direct answer to the sub-problem using its internal knowledge, as guided by the prompt detailed in Appendix~\ref{appendix.knowledge_boundary_determination}. Subsequently, the reliability of this initial response is evaluated via a hybrid confidence assessment that combines both prompt-based and likelihood-based methods. The final confidence is determined to be \textit{True} if and only if both methods independently yield a \textit{True} assessment. If the confidence level is deemed sufficient, this internally generated answer is adopted as the final solution.

\noindent\textbf{Prompt-based Confidence Assessment. } 
We introduce a prompt-based methodology for confidence assessment that employs introspective verification. This interactive procedure enables the model to leverage its internal knowledge for the iterative self-assessment of its generated answers, culminating in a binary classification of \textit{True} or \textit{False}. The prompt template utilized in this process is specified in Appendix~\ref{appendix.knowledge_boundary_determination}.

\begin{figure}[htbp]
    \centering
    \includegraphics[width=\linewidth]{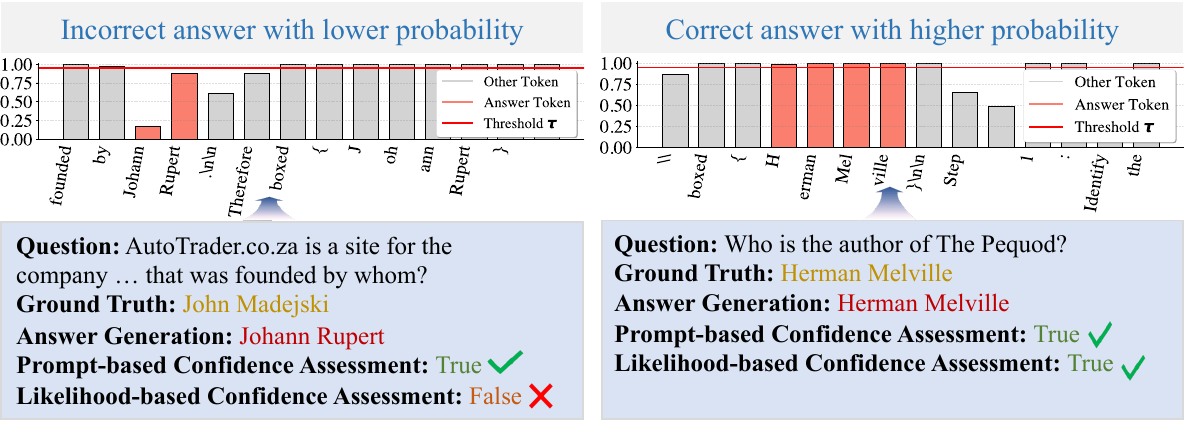}
    \caption{The probability distribution of the answer tokens. The left shows the low confidence answers, and the right shows the high confidence answers. }
    \label{fig:token_probs}
\end{figure}

\noindent\textbf{Likelihood-based Confidence Assessment. }
Likelihood-based confidence assessment evaluates the reliability of a model's generated response by leveraging the probabilities of its output tokens. As illustrated in Figure~\ref{fig:token_probs}, given a sub-problem input $\boldsymbol{x}$, the model first extracts the critical content enclosed within \textbackslash boxed\{\} from its response. This extracted content is formulated as a target sequence $\boldsymbol{y} = \{y_1, y_2, \dots, y_T\}$, where $T$ is the sequence length. During the auto-regressive generation process, each token $y_t$ ($t = 1, 2, \dots, T$) is associated with a conditional probability $P(y_t \mid \boldsymbol{x}, y_{<t})$. This term represents the likelihood of generating $y_t$ given the input $\boldsymbol{x}$ and the preceding context $y_{<t}$. While these token-level probabilities can, in principle, serve as a basis for confidence evaluation, a naive averaging of these values is an inadequate strategy. This is because high-frequency tokens (e.g., ``the'', ``of'') inherently possess high generation probabilities but contribute minimally to assessing the model's certainty regarding the core semantic content. To more effectively capture moments of high uncertainty during generation, we adopt a minimum probability criterion. We define the confidence score $C$ as the lowest probability among all tokens in the target sequence:
\begin{center}
$C = \min \{p(y_{1} \mid \boldsymbol{x}), p(y_{2} \mid \boldsymbol{y}_{<2}, \boldsymbol{x}), \dots, p(y_{T} \mid \boldsymbol{y}_{<T}, \boldsymbol{x})\}$
\end{center}
The reliability of a generated answer is assessed by comparing its confidence score, $C$ against a predefined threshold, $\tau$. An answer is classified as reliable (\textit{True}) if $C \geq \tau$, and unreliable (\textit{False}) otherwise.

\subsection{Focusing and Reasoning}
\label{sec:focusing_reasoning}
During knowledge boundary determination, we ascertain whether a sub-question requires retrieval. After executing retrieval, we need to assess if the current retrieval results can adequately address the sub-problem. Therefore, we have designed Focusing and Reasoning modules to fulfill this purpose. Focusing and Reasoning are primarily designed to analyze retrieved references, with their data construction methodology as detailed in Appendix~\ref{appendix.focusing_and_reasoning}. Subsequently, based on the answer and reason from the Output section of Appendix~\ref{appendix.focusing_and_reasoning}, we employ LLMs to perform conditional restructuring, ultimately embedding the restructured content into the designated $<$think$>$ within the depth-solving module.

\subsection{Multi-Turn Interactive Training}
We conduct supervised fine-tuning of the deep search reasoning process through multi-turn interaction data. Detailed descriptions of the complete training samples and the corresponding data construction methodology can be found in Appendix~\ref{appendix.training_sample} and Appendix~\ref{appendix:data_construction}, respectively.
The supervised fine-tuning procedure for multi-turn interactions closely follows that of the single-turn setting. Specifically, each entire interaction sequence, represented as $[S,U_1,A_1,...,U_n,A_n]$, is concatenated. Where $S_i$, $U_i$, and $A_i$ denote the System, User, and Assistant contents, respectively.
The cross-entropy loss is computed solely over the assistant’s response tokens ($A_1...A_n$), after which the losses are averaged. This approach is computationally efficient due to its compatibility with parallel processing, and it maximizes the utility of the data by incorporating supervision at every assistant turn.
This supervised learning strategy enables the model to acquire and adapt to specific response styles, thereby facilitating customization for specialized domains such as medicine, law, and finance. This capability constitutes a significant advantage over direct reinforcement learning rollouts, in which exerting control over response style is considerably more challenging.

\section{Experiments}
\renewcommand\arraystretch{1.0}
\begin{table*}[ht]
\centering
\small
\setlength\aboverulesep{0pt}\setlength\belowrulesep{0pt}
\begin{tabular}{c|l|ccc|cccc|c}
\toprule
\multirow{2}{*}{\textbf{LLMs}}                 & \multicolumn{1}{c|}{\multirow{2}{*}{\textbf{Methods}}} & \multicolumn{3}{c|}{\textbf{Single-Hop QA}}                & \multicolumn{4}{c|}{\textbf{Multi-Hop QA}}                                  & \multirow{2}{*}{\textbf{Avg}} \\ \cline{3-9}
                                      & \multicolumn{1}{c|}{}                         &  NQ$^\dagger$      & TriviaQA$^\ast$   & PopQA$^\ast$   & HotpotQA$^\dagger$ & 2Wiki$^\ast$ & MuSiQue$^\ast$ & Bamboogle$^\ast$ &                      \\ \midrule
\multirow{11}{*}{\rotatebox{90}{Qwen2.5-3B-Instruct}} & Naive Generation                             & 0.106          & 0.288          & 0.108          & 0.149          & 0.244          & 0.020          & 0.024          & 0.134                \\
                                      & CoT                                          & 0.023          & 0.032          & 0.005          & 0.021          & 0.021          & 0.002          & 0.000          & 0.015                \\ \cline{2-10}
                                      & Search-o1                                    & 0.238          & 0.472          & 0.262          & 0.221          & 0.218          & 0.054          & 0.320          & 0.255                \\
                                      & IRCoT                                        & 0.111          & 0.312          & 0.200          & 0.164          & 0.171          & 0.067          & 0.240          & 0.181                \\
                                      & Naive RAG                                    & 0.348          & 0.544          & 0.387          & 0.255          & 0.226          & 0.047          & 0.080          & 0.270                \\ \cline{2-10}
                                      & R1-Gen                                       & 0.210          & 0.449          & 0.171          & 0.208          & 0.275          & 0.06           & 0.192          & 0.224                \\
                                      & Search-R1                                    & 0.341          & 0.545          & 0.378          & 0.324          & 0.319          & 0.103          & 0.264          & 0.325                \\
                                      & ZeroSearch                                   & 0.414          & 0.574          & 0.448          & 0.274          & 0.300          & 0.098          & 0.111          & 0.317                \\ \cdashline{2-10}[2pt/2pt]
                                      & StepSearch                                   & -              & -              & -              & 0.345          & 0.320          & 0.174          & 0.344          & -                    \\
                                      & ReSearch                                     & 0.352          & 0.544          & 0.421          & 0.356          & 0.331          & 0.159          & 0.280          & 0.349                \\ \cline{2-10}
                                      & Thinker(ours)                                & \textbf{0.439} & \textbf{0.598} & \textbf{0.469} & \textbf{0.400}   & \textbf{0.469} & \textbf{0.214} & \textbf{0.424} & \textbf{0.430}       \\ \midrule
\multirow{11}{*}{\rotatebox{90}{Qwen2.5-7B-Instruct}} & Naive Generation                             & 0.134          & 0.408          & 0.140          & 0.183          & 0.250          & 0.031          & 0.120          & 0.181                \\
                                      & CoT                                          & 0.048          & 0.185          & 0.054          & 0.092          & 0.111          & 0.022          & 0.232          & 0.106                \\ \cline{2-10}
                                      & Search-o1                                    & 0.151          & 0.443          & 0.131          & 0.187          & 0.176          & 0.058          & 0.296          & 0.206                \\
                                      & IRCoT                                        & 0.224          & 0.478          & 0.301          & 0.133          & 0.149          & 0.072          & 0.224          & 0.226                \\
                                      & Naive RAG                                    & 0.349          & 0.585          & 0.392          & 0.299          & 0.235          & 0.058          & 0.208          & 0.304                \\ \cline{2-10}
                                      & R1-Gen                                       & 0.270          & 0.537          & 0.199          & 0.237          & 0.292          & 0.072          & 0.293          & 0.271                \\
                                      & Search-R1                                    & 0.393          & 0.610          & 0.397          & 0.370          & 0.414          & 0.146          & 0.368          & 0.385                \\
                                      & ZeroSearch                                   & 0.436          & \textbf{0.652} & \textbf{0.488} & 0.346          & 0.352          & 0.184          & 0.278          & 0.391                \\ \cdashline{2-10}[2pt/2pt]
                                      & StepSearch                                   & -              & -              & -              & 0.386          & 0.366          & 0.226          & 0.400          & -                    \\
                                      & ReSearch                                     & 0.407          & 0.611          & 0.423          & 0.419          & 0.412          & 0.205          & 0.400          & 0.411                \\ \cline{2-10}
                                      & Thinker(ours)                                & \textbf{0.450} & 0.642          & 0.484          & \textbf{0.421} & \textbf{0.469} & \textbf{0.221} & \textbf{0.480} & \textbf{0.452} \\ \bottomrule     
\end{tabular}
\caption{EM performance of different models. The best performance is set in bold. $^\dagger$/$^\ast$ represents in-domain / out-domain datasets. In contrast to other baselines, StepSearch and ReSearch employ the Musique dataset for training.}
\label{tab:main_results}
\end{table*}
\subsection{Experimental Settings}
\textbf{Benchmarks.} Our experiments are conducted on 7 widely-used datasets. The dataset comprises two primary categories: (1) Single-Hop QA: NQ~\cite{kwiatkowski-etal-2019-natural}, TriviaQA~\cite{joshi-etal-2017-triviaqa}, and PopQA~\cite{mallen-etal-2023-trust}. (2) Multi-Hop QA: HotpotQA~\cite{yang-etal-2018-hotpotqa}, 2WikiMultiHopQA~\cite{ho-etal-2020-constructing}, Musique~\cite{trivedi-etal-2022-musique}, and Bamboogle~\cite{press2023measuringnarrowingcompositionalitygap}. These datasets encompass diverse search and reasoning challenges, thereby enabling comprehensive evaluation of our model. Our evaluation set maintains methodological consistency with established prior works~\cite{chen2025researchlearningreasonsearch,jin2025searchr1trainingllmsreason}.

\noindent\textbf{Comparison Methods.}
We establish a three-tiered evaluation model comprising the following baselines:
(1) \textbf{Non-Retrieval Paradigms} Naive Generation: Direct answer synthesis without external knowledge integration; Chain-of-Thought (CoT): Explicit cognitive pathway formalization through sequential reasoning traces~\cite{cotWei2023}. (2) \textbf{Retrieval-Augmented Architectures} Naive RAG: Standard retrieval-generation pipeline without iterative optimization~\cite{DBLP:conf/nips/LewisPPPKGKLYR020}; IRCoT: Multi-cycle retrieval-reasoning coordination with dynamic feedback mechanisms~\cite{DBLP:conf/acl/TrivediBKS23}; Search-o1: Agent-mediated search workflow integration in reasoning processes~\cite{Search-o1}. (3) \textbf{Reinforcement Learning Models} R1-Gen: Pure RL-optimized generation without search engine interfacing~\cite{deepseekai2025deepseekr1incentivizingreasoningcapability}; Search-R1: Multimodal trajectory optimization combining search interactions with reasoning steps~\cite{jin2025searchr1trainingllmsreason}; ZeroSearch: Supervised LLM transformation into dual-function retrieval module (relevant/noisy document generation)~\cite{sun2025zerosearchincentivizesearchcapability}; StepSearch: PPO-based search LLM training with incremental exploration~\cite{wang2025stepsearchignitingllmssearch}. 

\noindent\textbf{Implementation Details.}
To maintain consistency with previous work, we selected NQ and HotpotQA as our training datasets. Using the data construction and evaluation frameworks outlined in Appendices~\ref{appendix:data_construction} and~\ref{appendix:data_evaluation}, we obtained a total of 71K training samples. All comparison methods use the same base model. We employ E5-base-v2~\cite{wang-etal-2024-improving-text} as the retriever and utilize Wikipedia data from December 2018 as the knowledge base~\cite{karpukhin-etal-2020-dense}. All corpus indexing and embedding processes are pre-processed using FlashRAG~\cite{flashrag2025}. Except for ReSearch, which retrieves the top 5 documents for each question, all other methods retrieve the top 3 documents. For additional implementation details, please refer to Appendix~\ref{appendix:implementation_details}. 



\subsection{Main Results}
We evaluated our model on seven widely used benchmark datasets. Here, to maintain consistency with the retriever components of other baseline methods, we conduct plain text retrieval using the \textit{``Step"} content from the logical form. Table~\ref{tab:main_results} presents the comparisons between our model and other baselines. 
As the results show, our model significantly outperforms the baselines at both the 3B and 7B scales. The improvement is particularly notable for the 3B model, which achieves an average 7.9\% gain over the Research baseline. This is because the 3B model's low-quality RL rollouts struggled to produce correct answers. For experiments involving larger models and different base models, please refer to Appendix~\ref{appendix.14b_llm} and ~\ref{appendix.different_llms}. We take the 7B model as an example for a detailed analysis.
Compared to baselines without retrieval, the performance of our model exceeds Naive Generation and CoT by an average of 27. 1\% and 34. 6\%, respectively, in seven datasets. The primary reason for the improvement is that, within these datasets, retrieval is essential, as LLMs struggle to directly answer these questions accurately without access to relevant information.
Compared to retrieval-augmented methods, our model outperforms Search-o1, IRCoT, and Naive RAG by average margins of 24.6\%, 22.6\%, and 14.8\%, respectively. These enhancements are primarily attributed to our model's implementation of breadth decomposition and depth solving, enabling it to learn how to leverage the retriever more effectively—particularly during deep retrieval, where the generated queries align well with the retriever's capabilities.
In comparison to reinforcement learning-based approaches, it can be observed that our model outperforms the previous state-of-the-art model, ReSearch, by an average of 4.1\% in EM score across seven datasets. Specifically, it improves by an average of 4.5\% on single-hop datasets and by an average of 3.9\% on multi-hop datasets. The primary reason is that our breadth decomposition and depth solving model enables the decomposition of questions into atomic granularity, thereby reducing the complexity involved in model retrieval and answering.

The logical coherence and rigor of different deep search models are evaluated using four defined metrics: Logical Hierarchy (Hier), Interleaved Solving (Intrlv), Granularity Consistency (Gran), and Search Efficiency (Eff) (see Appendix~\ref{append.logical_metrics} for detailed definitions). For this evaluation, GPT-4 is employed to assess the reasoning processes on the HotpotQA test set. The results, presented in Table~\ref{tab.logical_eval}, indicate that the performance of Thinker in logical coherence and rigor is significantly superior to that of ReSearch and Search-R1 across these dimensions.
\renewcommand\arraystretch{1.0}
\begin{table}[htbp]
\small
\centering
\setlength\aboverulesep{0pt}\setlength\belowrulesep{0pt}\setlength{\tabcolsep}{7pt}
\begin{tabular}{l|ccccc}
\toprule
\textbf{Methods} & \textbf{Hier} & \textbf{Intrlv} & \textbf{Gran} & \textbf{Eff} & \textbf{Overall}   \\ \midrule
Search-R1 & 0.813     & 0.955        & 0.852       & 0.903      & 0.638 \\
ReSearch  & 0.872     & 0.967        & 0.877       & 0.922      & 0.705 \\
Thinker   & \textbf{0.975}     & \textbf{0.989}        & \textbf{0.955}       & \textbf{0.958}      & \textbf{0.904} \\ \bottomrule 
\end{tabular}
\caption{The accuracy of logical coherence and rigor across different models on HotpotQA.}
\label{tab.logical_eval}
\end{table}

\subsection{Model Ablation Studies}
To evaluate the contribution of each component within Thinker (Qwen2.5-7B-Instruct), we conduct an ablation study across seven datasets (cf. Table~\ref{tab.model_ablation}). The removal of the depth solving component causes the most significant performance degradation, resulting in a 3.7\% drop in the average EM score. This is primarily because depth solving retrieves sufficient external knowledge for each sub-problem, which is crucial for improving its solution accuracy. In contrast, ablating the focusing and reasoning module results in a slight performance decrease. Furthermore, while removing the knowledge boundary determination has a minimal impact on overall performance, it significantly reduces unnecessary retrievals (cf. Table~\ref{tab.knowledge_boundary_determination}). A similar observation applies to the logical function (the Action part of a sub-problem): although its contribution to overall performance is negligible, its role is indispensable for enhancing performance within the graph-retrieval-enabled framework.
\renewcommand\arraystretch{1.0}
\begin{table}[ht]
\small
\centering
\setlength\aboverulesep{0pt}\setlength\belowrulesep{0pt}\setlength{\tabcolsep}{5pt}
\begin{tabular}{l|c}
\toprule
                                 & \textbf{Avg}   \\  \midrule
Thinker                          & 0.452 \\
\quad - depth solving                    & 0.415 \\
\quad - knowledge boundary determination & 0.447 \\
\quad - focusing \& reasoning            & 0.445 \\
\quad - logical function                 & 0.449 \\ \bottomrule
\end{tabular}
\caption{ Model ablation studies (average EM score).}
\label{tab.model_ablation}
\end{table}

\subsection{Sensitivity Analysis}
\label{sec.sensitivity_analysis}
To validate the robustness of Thinker (Qwen2.5-7B-Instruct), we conduct a series of sensitivity analysis experiments.
As shown in Table~\ref{tab.data_ratio}, we evaluate the performance of Thinker with varying numbers of training samples. We observe that even with only 1\% of the data, which corresponds to just a few hundred samples, the performance is already close to SOTA method (ReSearch Avg: 0.411). As the sample size increases, the performance also shows a gradual improvement. This also demonstrates that our multi-turn interaction approach significantly reduces training costs.
\renewcommand\arraystretch{1.0}
\begin{table}[htbp]
\small
\centering
\setlength\aboverulesep{0pt}\setlength\belowrulesep{0pt}\setlength{\tabcolsep}{5pt}
\begin{tabular}{l|cccccc}
\toprule
\textbf{Dataset} & \textbf{1\%}  & \textbf{10\%} & \textbf{20\%}  & \textbf{50\%}  & \textbf{80\%}  & \textbf{100\%} \\ \midrule
NQ        & 0.388 & 0.441 & 0.439 & 0.453 & 0.452 & 0.450 \\
TriviaQA  & 0.602 & 0.614 & 0.620 & 0.627 & 0.625 & 0.642 \\
PopQA     & 0.438 & 0.469 & 0.472 & 0.478 & 0.475 & 0.484 \\
HotpotQA  & 0.377 & 0.408 & 0.412 & 0.414 & 0.412 & 0.421 \\
2Wiki     & 0.413 & 0.438 & 0.461 & 0.471 & 0.470 & 0.469 \\
MuSiQue   & 0.200 & 0.212 & 0.214 & 0.216 & 0.221 & 0.221 \\
Bamboogle & 0.424 & 0.432 & 0.448 & 0.472 & 0.456 & 0.480 \\ \hline
Avg       & \textbf{0.406} & 0.431 & 0.438 & 0.447 & 0.444 & 0.452 \\ \bottomrule
\end{tabular}
\caption{Impact of the number of training samples.}
\label{tab.data_ratio}
\end{table}

Table~\ref{tab.max_retrievals} illustrates the impact of the maximum search depth per sub-problem on EM performance. The results indicate a significant performance degradation when the model is limited to a single search. In contrast, the performance stabilizes once the maximum search count is two or greater. Notably, we find that single-hop datasets are insensitive to retrieval depth. This is primarily because the sub-problems in these tasks are relatively straightforward, allowing the majority of answers to be retrieved directly.
\renewcommand\arraystretch{1.0}
\begin{table}[htbp]
\small
\centering
\setlength\aboverulesep{0pt}\setlength\belowrulesep{0pt}\setlength{\tabcolsep}{7.5pt}
\begin{tabular}{l|cccccc}
\toprule
\textbf{Dataset} & \textbf{D=1}  & \textbf{D=2}  & \textbf{D=3}   & \textbf{D=4}   & \textbf{D=5}   \\ \midrule
NQ        & 0.427 & 0.438 & 0.437 & 0.440 & 0.450 \\
TriviaQA  & 0.627 & 0.632 & 0.627 & 0.630 & 0.642 \\
PopQA     & 0.464 & 0.470 & 0.470 & 0.470 & 0.484 \\
HotpotQA  & 0.375 & 0.420 & 0.410 & 0.408 & 0.421 \\
2Wiki     & 0.421 & 0.456 & 0.460 & 0.455 & 0.469 \\
MuSiQue   & 0.181 & 0.230 & 0.225 & 0.225 & 0.221 \\
Bamboogle & 0.408 & 0.448 & 0.464 & 0.448 & 0.480 \\ \hline
Avg       & 0.415 & 0.442 & 0.442 & 0.439 & 0.452 \\ \bottomrule
\end{tabular}
\caption{Impact of the maximum number of depth searches.}
\label{tab.max_retrievals}
\end{table}

\renewcommand\arraystretch{1.0}
\begin{table*}[ht]
    \small
    \centering
\setlength\aboverulesep{0pt}\setlength\belowrulesep{0pt}
\begin{tabular}{l|c|l|cc|cc|cc|cc}
\toprule
\multirow{2}{*}{\textbf{Methods}}   & 
\multirow{2}{*}{\textbf{Size}}    & 
\multirow{2}{*}{\textbf{Retriever}}    & 
\multicolumn{2}{c|}{\textbf{HotpotQA}} &    
\multicolumn{2}{c|}{\textbf{MusiQue}} &   
\multicolumn{2}{c|}{\textbf{2Wiki}}   & 
\multicolumn{2}{c}{\textbf{Avg}}  \\ \cline{4-11}    
            &       &  & EM       & F1     & EM      & F1    & EM    & F1  & EM    & F1    \\
            \midrule
Search-R1 & 7B & BGE-M3
& 0.545    & 0.676 & 0.307   & 0.403  & 0.517 & 0.589  & 0.456 & 0.556  \\
ReSearch & 7B & BGE-M3 & 0.538    & 0.671 &  0.322   & 0.423   & 0.555  & 0.633   & 0.472  & 0.576 \\
Thinker (ours)  & 7B & BGE-M3 & 0.538    & 0.664 &    0.337    & 0.442    & 0.596 & 0.657   & 0.490 & 0.588  \\ \hline
KAG-Thinker (ours) & 7B  & Hybrid Graph Retriever
& \textbf{0.568} & \textbf{0.698} & \textbf{0.350}   & \textbf{0.469} & \textbf{0.644} & \textbf{0.711}  & \textbf{0.520} & \textbf{0.626} \\
\bottomrule
\end{tabular}
    \caption{Performance of different models within a self-built corpus. All retrievers retrieve the top 3 documents.}
    \label{tab:retriever_experiments}
\end{table*}
\begin{figure*}[ht]
\centering 
\includegraphics[width=\textwidth]{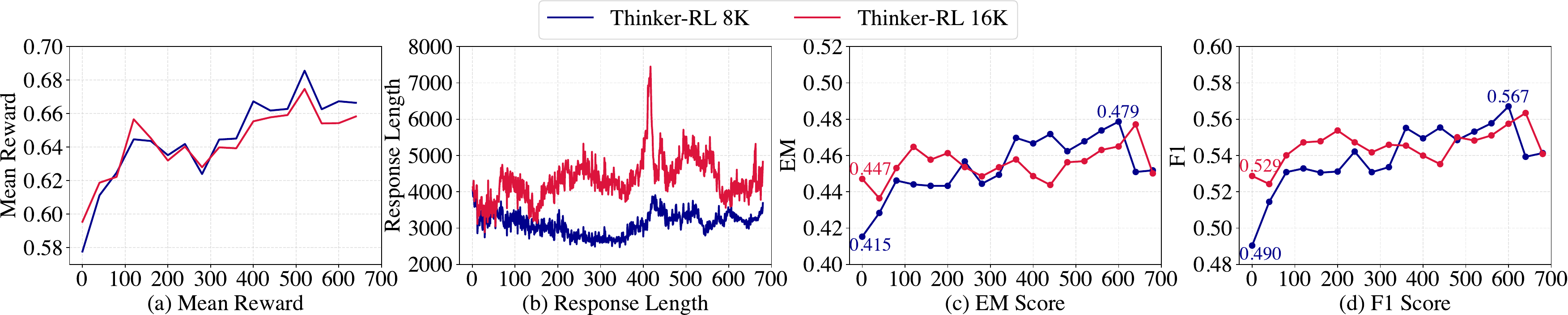} 
\caption{Performance of Thinker with Reinforcement Learning vs. Training Steps. (a) Average reward during the RL training. (b) Average model response length. (c) Average EM score across seven datasets. (d) Average F1 score across seven datasets.} 
\label{Fig.Thinker_RL} 
\end{figure*}

To evaluate the performance of the knowledge boundary determination, we conduct an analysis on the test sets of TriviaQA and Bamboogle, with the results presented in Table~\ref{tab.knowledge_boundary_determination}. In our evaluation, EM=1 signifies a correct prediction, whereas EM=0 signifies an incorrect one. In cases where EM=1, we assume that all decisions not to retrieve are accurate, as the LLM successfully answers the sub-question using its internal knowledge, which results in an overall accuracy of 1 for this subset. Conversely, in the EM=0 samples, there is a notable reduction in sub-questions for which retrieval is deemed unnecessary. In summary, the integration of the knowledge boundary determination module leads to 16.0\% and 17.8\% reductions in search queries on TriviaQA and Bamboogle, respectively. However, for instances where EM=0, assessing the correctness of a no-retrieval determination is challenging. To calculate its accuracy under this condition, we employ a dual verification process using Qwen2.5-72B-Instruct and DeepSeek-R1-Distill-32B. A no-retrieval determination is considered correct only if both models concur that the answer generated by Thinker based on its internal knowledge is accurate. As shown in Table~\ref{tab.knowledge_boundary_determination}, our knowledge boundary determination achieves an accuracy of 96.9\% on TriviaQA and 97.8\% on Bamboogle.
\renewcommand\arraystretch{1.0}
\begin{table}[htbp]
\small
\centering
\setlength\aboverulesep{0pt}\setlength\belowrulesep{0pt}\setlength{\tabcolsep}{5pt}
\begin{tabular}{l|cc|cc|cc}
\toprule
\multirow{2}{*}{\textbf{Dataset}} & \multicolumn{2}{c|}{\textbf{EM = 1}} & \multicolumn{2}{c|}{\textbf{EM = 0}} & \multicolumn{2}{c}{\textbf{All}} \\ \cline{2-7}
                         & KBD          & Acc         & KBD          & Acc         & KBD        & Acc        \\ \midrule
TriviaQA                 & 0.216        & 1.000       & 0.065        & 0.794       & 0.160      & 0.969      \\
Bamboogle                & 0.234        & 1.000       & 0.124        & 0.938       & 0.178      & 0.978      \\ \bottomrule
\end{tabular}
\caption{Retrieval rate reduction from knowledge boundary determination (KBD).}
\label{tab.knowledge_boundary_determination}
\end{table}

\subsection{Thinker with KAG}
Our dual representation of sub-problems enables better utilization of high-quality knowledge bases. To validate this hypothesis, we create KAG-Thinker by applying the Thinker model within the KAG framework~\cite{DBLP:journals/corr/abs-2409-13731}. The multi-turn interactive reasoning framework of KAG-Thinker is depicted in Appendix~\ref{appendix.thinker_with_kag}.
To maintain consistency with KAG, we continue to use its test set, with 1,000 examples each from HotpotQA, Musique, and 2Wiki (consistent with HippoRAG~\cite{DBLP:conf/nips/GutierrezS0Y024}). The data for our self-built corpus are entirely derived from the supporting facts of these three datasets, totaling 26,990 documents. Within the self-built corpus, and utilizing the same retriever and retrieved documents, our Thinker model still outperforms ReSearch and Search-R1 across three multi-hop datasets.
For a fair comparison against the baselines, the original Thinker model could only perform pure text retrieval with BGE-M3 for natural language steps in its logical form, even for operations like Deduce and Math. However, the KAG framework robustly supports these operations. Table~\ref{tab:retriever_experiments} illustrates that KAG-Thinker, operating within the KAG framework, shows a marked improvement over Thinker, with average EM and F1 scores increasing by 3.0\% and 3.8\%, respectively. KAG-Thinker achieves these enhancements by leveraging two key features from the KAG framework: the hybrid graph retriever (HGR) and native support for Math and Deduce. 

\subsection{Thinker with Reinforcement Learning}
To explore the potential of the Thinker model, we apply reinforcement learning initialized from the optimal SFT model. For implementation details, please see Appendix~\ref{appendix:implementation_details}. As illustrated in Figure~\ref{Fig.Thinker_RL}, our experiments show significant improvements. Figure~\ref{Fig.Thinker_RL}(a) plots the training reward curve, smoothed with a 40-step moving average, while Figure~\ref{Fig.Thinker_RL}(b) tracks the average response length. Figure~\ref{Fig.Thinker_RL}(c) and (d) show the average EM and F1 scores on seven evaluation datasets. The application of reinforcement learning yields a substantial performance gain over the SFT baseline, increasing the average EM score from 0.452 to 0.479. For a detailed performance comparison on each dataset, please refer to Appendix~\ref{appendix.thinker_rl}. This improvement is strongly correlated with the training reward, demonstrating RL's ability to unlock latent model potential. Moreover, experiments with output length constraints show that a more restrictive 8K limit, despite initial underperformance, ultimately outperforms the 16K counterpart, effectively promoting response conciseness. Compared to deep search methods based on a pure reinforcement learning algorithm, our SFT-then-RL methodology not only preserves the logical consistency of reasoning but also significantly boosts the model's performance. Ultimately, Thinker-RL performs inference in the same way as Thinker.

\section{Conclusion}
In this paper, we propose a model for interactive thinking and deep reasoning to solve complex problems. We find that our model significantly outperforms previous methods across various datasets and model sizes, and notably improves the logical rigor in solving complex tasks. Through ablation studies and sensitivity analyses, we discover that our proposed methods can achieve near-SOTA performance even when trained on only a few hundred samples. At the same time, it reduces unnecessary retrievals by using knowledge boundary determination. Our approach offers a systematic reasoning framework, making it particularly well-suited for domains where logical rigor is paramount, including medical diagnostics (cf. Appendix~\ref{appendix.medical_application}), legal compliance, and financial risk assessment.

\bibliography{aaai2026}

\clearpage
\appendix

\section{Data Construction Framework}
\label{appendix:data_construction}
\begin{figure}[b!]
\centering 
\includegraphics[width=0.45
\textwidth]{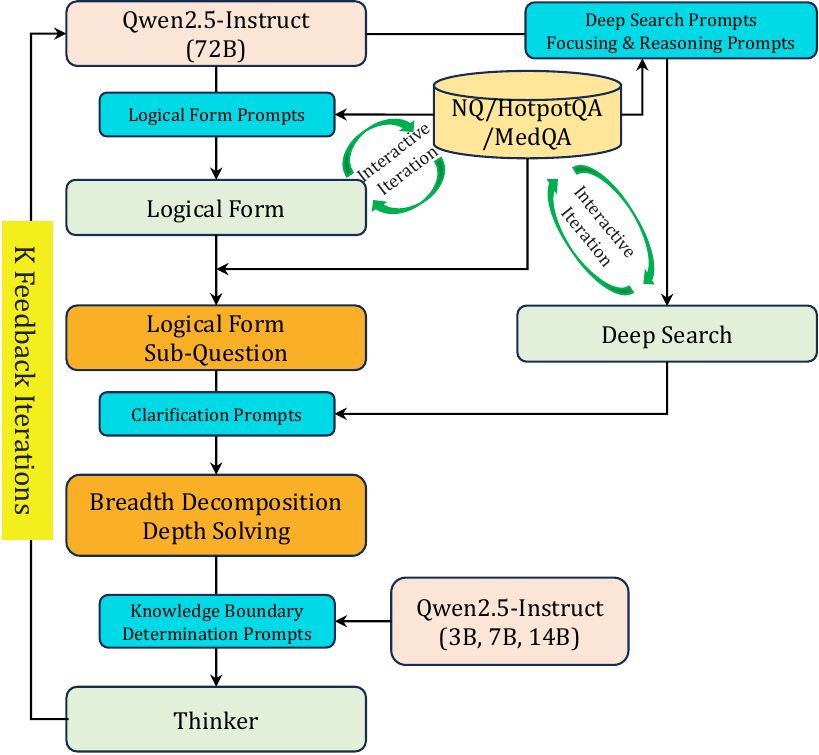} 
\caption{Overview of the data construction framework.} 
\label{Fig.dataset_construction} 
\end{figure}
RL-based methods often struggle to control their outputs in a structured manner. They cannot effectively customize the intermediate reasoning steps during decision-making. Instead, these methods rely on random exploration within LLMs, leading to unpredictable reasoning paths. This unpredictability is particularly problematic in critical professional fields that require transparent and auditable processes, such as medical diagnostics, legal compliance, and financial risk assessment. To address these limitations, we have developed a systematic Search-QA synthesis framework with a customizable reasoning process, as depicted in Figure~\ref{Fig.dataset_construction}. Our data synthesis primarily covers the general and medical domains. For the general domain, the data construction process involves three main stages: 

\noindent\textbf{Logical Form Data Generation}: We first design a series of logical form prompts (Appendix~\ref{appendix:logical_form}) and use Qwen2.5-72B-Instruct to perform inference on the NQ and HotpotQA datasets. This generates the initial "natural language to logical form" training data. After filtering for correctness with our data evaluation framework (Appendix~\ref{appendix:data_evaluation}) and undergoing multiple rounds of iteration, we obtain the final Logical Form model training set, comprising 59K samples.

\noindent\textbf{Deep Search Data Generation}: In parallel, we design prompts for deep search and focused reasoning. We perform inference on single-hop questions from the NQ and HotpotQA datasets using Qwen2.5-72B-Instruct to generate data for a sub-question deep search model. After several iterative refinements, we compile the final Deep Search model training set, which consists of 18K samples.

\noindent\textbf{Thinker Data Synthesis}: Finally, leveraging the Logical Form and Deep Search models, we generate a breadth-first decomposition and a deep-solving path for each question. For each sub-question, we perform knowledge boundary determination: if an LLM can confidently answer it, we omit its deep-solving process to reduce unnecessary retrievals. Through multiple cycles of iteration and validation with the data evaluation framework, we construct the final Thinker training dataset, containing 71K samples.

\section{Logical Form}
\label{appendix:logical_form}
Logical Functions are defined in Table~\ref{tab:function-syntax}. Each function represents an execution action. Complex problems are decomposed by planning a combination of these expressions, enabling reasoning about intricate issues.
\renewcommand\arraystretch{1.1}
\begin{table*}[ht!]
    \small
    \centering
    \setlength\aboverulesep{0pt}\setlength\belowrulesep{0pt}
    \begin{tabular}[c]{p{1.4 cm}|p{7 cm}|p{6.5 cm}}
       \toprule
       \textbf{Function} & \textbf{Function Expression} & \textbf{Description} \\
       \cline{1-3} 
       \midrule
       \textit{Retrieval}  
       & $Retrieval$($s$ = $s_i$:type[name], $p$ = $p_i$:edge, $o$ = $o_i$:type[name], $s$.prop = value, $p$.prop = value, $o$.prop = value) & Retrieve $<s, p, o>$ triples. Conditions can be set as constraints, and $s_i, p_i, o_i$ serve as variable names for reference in subsequent action planning. When referring to previously mentioned variable names, there is no need to regenerate entity types and names. \\
       \midrule
       \textit{Deduce} 
       & $Deduce$(\texttt{op} = extract|judgement|entailment|choice,        \texttt{content} = [A,B,...], \texttt{target} = [...]) $\rightarrow$ $deduce_i$ &  The parameter \texttt{op} can specify different inference tasks, such as information extraction, judgment, logical reasoning, single-choice, multiple-choice, etc. The parameter \texttt{content} specifies contextual information and can be either texts or variables. The \texttt{target} specifies the objective of the inference. \\
       \midrule
       \textit{Math} & $Math$ (\texttt{content} = [A,B,...], \texttt{target} = [...]) $\rightarrow$ $math_i$ &  Numerical or statistical calculations can be performed. The \texttt{content} specifies contextual information and can be either text or variables. The \texttt{target} specifies the calculation objective. \\
       \midrule
       \textit{Output} & $Output$ (A,B,...) & Directly output A, B, ... as the answer, where A and B are variable names referring to the results of previous retrieval, deduce, or calculation. \\
       \bottomrule
    \end{tabular}
    \caption{Different functions of logical form.}
    \label{tab:function-syntax}
\end{table*}

\noindent\textbf{Retrieval.} The logical function of Retrieval is defined as $Retrieval$(\textit{$s$, $p$, $o$, $s$.constraints, $p$.constraints, $o$.constraints}). The function's variables consist of a Subject-Predicate-Object (SPO) ($s$, $p$, $o$), along with conditional constraints that can be applied to any of the three elements. For example, the question \textit{``Who was the director of the 2002 film Men in Black?''} can be transformed into $Retrieval$ (\textit{$s$: Movie[Men in Black], $p$: DirectedBy, $o$: Person, $s$.ReleaseYear=2002}). The final retrieval results include the subgraph $SG(s, p, o)$ and chunks related to the target object $o$. 

\noindent\textbf{Deduce.} The logical function of Deduce is defined as $Deduce$ (\texttt{op}, \texttt{content}, \texttt{target}) $\rightarrow y_{d}$, where \texttt{op} specifies the reasoning method, including \textit{extract}, \textit{judgement}, \textit{entailment}, and \textit{choice}. The \texttt{content} field specifies the reasoning context, which includes the information related to the current sub-problem extracted from the user question and the calculation result of the previous logical forms passed through the intermediate variable. The \texttt{target} represents the current objective of the Deduce operation, and $y_{d}$ represents the result variable of the Deduce reasoning.

\noindent\textbf{Math.} The logical function of Math is defined as $Math$ (\texttt{content}, \texttt{target}) $\rightarrow y_{m}$. In this function, the \texttt{content} parameter specifies the known conditions required for the calculation, and the \texttt{target} parameter specifies the computation task. These parameters function similarly to their counterparts in the Deduce operation. $y_{m}$ represents the result of the computation.

\noindent\textbf{Output.} When the question can be answered using either the model's internal knowledge or the output of previous executors, the output executor will then determine the format and deliver the final answer.
\renewcommand\arraystretch{1.1}
\begin{table*}[tb!]
    \small
    \centering
    \setlength\aboverulesep{0pt}\setlength\belowrulesep{0pt}
    \begin{tabular}[l]{p{4.4 cm}|p{6.6 cm}|p{4.2 cm}}
        \toprule
         \multicolumn{3}{p{15.6 cm}}{ {\textbf{Question:} Zhang found a wallet on the street, which contained 5,000 yuan in cash and a credit card. Zhang took the cash and used  the credit card to make purchases at four stores: A, B, C, and D, for 210.4 yuan, 569.2 yuan, 1035.2 yuan, 2044.5 yuan, and 1035 yuan, respectiely. After the incident, the public security authorities determined that Zhang was suspected of embezzlement and credit card fraud. What is the total amount involved in the credit card fraud charge against Zhang?  } } \\
         
       \midrule
       \textbf{Steps} & \textbf{Actions}  & \textbf{Result} \\
       \cline{1-3} 
       \midrule
       1. What's the elements of the offense credit card fraud? 
       & Retrieval(s=s1:offense[credit card fraud], p=p1: found, o=o1:elements) & o1 = “The elements of the offense of credit card fraud include the following points: 1….” \\
       \midrule
       2. Which amounts are considered part of the credit card fraud amount?
       & Deduce(op=extract, content=[o1, Zhang found a wallet on the street…], target=Which amounts are considered part of the credit card fraud amount?)$\to$o2
  &  o2 = “The total amount involved in Zhang’s credit card fraud is 210. 4 yuan, 569.2 yuan, 1035.3 yuan, 2044.5 yuan, and 1035 yuan.”  \\
       \midrule
       3. What's the total amount involved in Zhang’s credit card fraud? & Math(content=[o1, o2], target=What's the total amount involved in Zhang’s credit card fraud.)$\to$o3 &   o3 = “the total amount involved in Zhang’s credit card fraud is 4989.40 yuan.” \\
       \midrule
      
       4. Output \#3.  & Output(o3)
   & Amount: 4894.40 yuan. \\
       \bottomrule
    \end{tabular}
    \caption{An example of the working principle of the Logical Form Solver in the integrated application of four executors:
    It search and references relevant legal provisions by \textbf{Retrieval}, extracts the amount of each illegal use of the credit card in the question by \textbf{Deduce}, calculates the amount involved in the credit card fraud case by \textbf{Math} and finally \textbf{Output} the answer.}
    \label{tab:solver example}
\end{table*}

Table \ref{tab:solver example} shows an example of using these four executors to solve the problem of \textit{calculating the amount of credit card fraud crime}. During the inference phase, Thinker generates equivalent representations of natural language and logical function for each logical form (such as Step1 and Action1 in table \ref{tab:solver example}), and then calls the corresponding Executor based on the dependencies.

\section{Data Evaluation Framework}
\label{appendix:data_evaluation}
Our training data is generated entirely by LLMs without any human involvement. However, the quality of this data directly impacts the model's final performance. To select high-quality data, we have designed a data evaluation framework.
The Data Evaluation Framework (DEF) constitutes a systematic methodology for optimal dataset curation through multi-dimensional quality assessment. In contemporary machine learning paradigms, empirical evidence consistently demonstrates that data quality exerts a more critical influence on model performance than dataset scale-a particularly salient consideration given the pervasive challenge of quality inconsistencies in publicly available corpora. Motivated by this critical need, we have engineered an evaluative architecture capable of granular quality diagnostics at the query-answer pair level. The DEF operates through six hierarchically organized assessment dimensions: security, accuracy, relevance, logic, fluency, and emotion.

\noindent\textbf{Security.} A security assessment identifies issues within data to ensure its safety and integrity. It checks if model outputs align with social norms, laws, and ethical standards. Key areas of focus include: (1) Content: Checking for sensitive, discriminatory, or inappropriate material to prevent harm.
(2) Privacy \& Security: Ensuring compliance with privacy laws and data security rules, such as preventing unauthorized access.
(3) Risk: Identifying and removing statements that could cause controversy or significant risks, which helps maintain trust and reputation.
(4) Consequences: Evaluating if actions based on the data might lead to negative outcomes.
In essence, this systematic check ensures data is accurate, legal, and ethical. It finds and resolves potential threats, fostering trust and security.

\noindent\textbf{Accuracy.}
Accuracy verification is a key process that identifies errors and inconsistencies in data and model outputs. It ensures model responses accurately address user questions and are factually sound. Key aspects of this evaluation include:
(1) Factual Error Detection: We rigorously check for any incorrect information, such as wrong dates, data, or logical flaws, to ensure reliability.
(2) User Intent Understanding: We verify that the model correctly interprets the user's question and its context, preventing irrelevant or misleading answers.
(3) Common Sense \& Domain Alignment: Outputs are checked against established knowledge and common sense principles for coherence and practical value.
(4) Illusion Detection: We look for answers that might seem correct but could actually mislead the user.
(5) Avoiding Over-Reasoning: We ensure the model doesn't provide excessive details or go beyond the scope of the question, keeping responses focused.
(6) Rejection Criteria: We define when an answer must be rejected due to irrelevance, inaccuracy, or confusion.
(7) Process Review: We examine the entire process of generating an answer for any factual errors introduced at any stage.
Crucially, judgments about factual correctness are based on the common sense embedded in the question itself. While external information can be helpful, the primary focus is on the question's inherent context and the user's expectations.

\noindent\textbf{Relevance.}
Relevance assessment evaluates whether model outputs are on-topic and appropriate for the given context. This process primarily checks how well the model's responses align with the input questions or cues. Key aspects of this evaluation include:
(1) Topic Deviation: We identify if the output strays from the main topic or includes information not directly relevant to the user's query. This ensures the information stays focused.
(2) Core Issue Identification: We assess the model's ability to pinpoint the main point of a question and provide specific answers that directly address it.
(3) Clarity and Conciseness: We ensure responses are not vague, too general, or overly detailed, which can obscure clarity. The goal is to provide concise, useful information without unnecessary embellishment.
In short, a relevance check is a systematic way to ensure model outputs are coherent and contextually appropriate. By identifying and fixing irrelevant content, we guarantee the information is not only accurate but also directly pertinent to the user's specific needs. This significantly improves the system's effectiveness and user satisfaction.

\noindent\textbf{Logic.}
Logical integrity assessment ensures that data outputs are internally consistent and free from errors. It verifies that model responses are rational and do not contain contradictions. Key areas of focus include:
(1) Logical Flow: Checking that timelines, sequences, and cause-and-effect relationships within the output are sound and consistent.
(2) Reasoning Soundness: Identifying any gaps in logical reasoning, flawed arguments, or incorrect deductions that undermine the conclusions.
(3) Real-World Alignment: Verifying that outputs align with established facts, common sense, and scientific principles governing real-world phenomena.
Ultimately, this meticulous process identifies and corrects logical flaws, thereby strengthening the reliability of the information and fostering confidence in data-driven decisions.

\noindent\textbf{Fluency.}
This assessment evaluates the language quality of model-generated text, focusing on its naturalness, coherence, and readability. Key aspects examined include:
(1) Structural Clarity and Grammar: We verify clear sentence structures and grammatical correctness to ensure easy comprehension.
(2) Conciseness: We aim to eliminate repetition, wordiness, and overly complex phrasing for clearer and more direct communication.
(3) Idiomatic Usage and Style: We ensure the language sounds natural, adhering to human-like idioms and appropriate stylistic conventions.
In essence, this assessment maintains high standards for linguistic excellence in model outputs. By systematically identifying and correcting fluency issues, we enhance overall text quality and readability, fostering effective communication and building trust.

\noindent\textbf{Emotion.}
This assessment checks the emotional and stylistic tone of model-generated outputs. Its main goal is to ensure the tone is appropriate for the given context and user. Key evaluation points include:
(1) Matching User Expectations: We ensure the output's tone (e.g., formal, friendly, humorous) aligns with what the user anticipates or what the situation demands.
(2) Avoiding Inappropriate Expressions: We prevent the model from using blunt, indifferent, or offensive tones that could alienate or confuse users.
(3) Contextual Adaptability: We verify the model can adjust its tone based on different contexts and various user preferences.
In short, this assessment ensures the model communicates with emotional intelligence and the right tone. By identifying and correcting tonal issues, it fosters more meaningful interactions, improves user engagement, and builds trust in the communication.

\renewcommand\arraystretch{1.1}
\begin{table}[htbp]
\centering
\small
\setlength\aboverulesep{0pt}\setlength\belowrulesep{0pt}
\begin{tabular}{p{1.2cm}|p{0.7cm}<{\centering}p{0.7cm}<{\centering}p{0.7cm}<{\centering}|p{0.7cm}<{\centering}p{0.7cm}<{\centering}p{0.7cm}<{\centering}}
\toprule
          & \multicolumn{3}{c|}{\textbf{Medical}}    & \multicolumn{3}{c}{\textbf{Legal}}        \\ \cline{2-7}
          & Good       & Bad       & Acc   & Good       & Bad       & Acc   \\ \hline
Security  & 72         & 0         & 1.000 & 96         & 0         & 1.000 \\
Accuracy  & 66         & 6         & 0.917 & 91         & 5         & 0.948 \\
Relevance & 70         & 2         & 0.972 & 94         & 2         & 0.979 \\
Logic     & 70         & 2         & 0.972 & 96         & 0         & 1.000 \\
Fluency   & 71         & 1         & 0.986 & 96         & 0         & 1.000 \\
Emotion   & 70         & 2         & 0.972 & 95         & 1         & 0.990 \\ 
All       & 64         & 8         & 0.888 & 90         & 6         & 0.938 \\ \bottomrule
\end{tabular}
\caption{Performance evaluation of a data evaluation framework in medical and legal domains.}
\label{tab:def_pos}
\end{table}
\renewcommand\arraystretch{1.1}
\begin{table}[htbp]
\centering
\small
\setlength\aboverulesep{0pt}\setlength\belowrulesep{0pt}
\begin{tabular}{p{1.2cm}|p{0.7cm}<{\centering}p{0.7cm}<{\centering}p{0.7cm}<{\centering}|p{0.7cm}<{\centering}p{0.7cm}<{\centering}p{0.7cm}<{\centering}}
\toprule
          & \multicolumn{3}{c|}{\textbf{Medical}}    & \multicolumn{3}{c}{\textbf{Legal}}        \\ \cline{2-7}
          & Good  & Bad  & Acc   & Good  & Bad  & Acc   \\ \hline
Security  & 24         & 5         & 0.828 & 36         & 9         & 0.800  \\
Accuracy  & 27         & 2         & 0.931 & 58         & 8         & 0.879  \\
Relevance & 24         & 6         & 0.800 & 42         & 8         & 0.840  \\
Logic     & 25         & 6         & 0.806 & 43         & 7         & 0.860  \\
Fluency   & 24         & 5         & 0.828 & 50         & 4         & 0.926  \\
Emotion   & 31         & 4         & 0.886 & 46         & 8         & 0.852  \\ \bottomrule
\end{tabular}
\caption{The data evaluation framework's performance in negative cases in medical and legal fields.}
\label{tab:def_neg}
\end{table}

To validate the performance of our data evaluation framework, we select positive and negative samples from the medical and legal domains for human evaluation.
The evaluation metrics for positive samples are systematically presented in Table~\ref{tab:def_pos}. Our DEF demonstrates robust performance across individual assessment dimensions, achieving accuracy rates exceeding 90\% in all criteria. Domain-specific composite evaluations yield 88.8\% accuracy in medical and 93.8\% in legal scenarios. Crucially, the framework exhibits controlled false positive rates during positive sample validation, indicating effective preservation of valid instances. The evaluation metrics for negative samples are comprehensively detailed in Table~\ref{tab:def_neg}. Our DEF achieves classification accuracy surpassing 80\% across all individual assessment criteria, with domain-specific composite accuracy reaching 88.6\% in medical and 85.2\% in legal. Through continuous optimization of the DEF, the quality of our data has been steadily improving.

\section{Implementation Details}
\label{appendix:implementation_details}
For the SFT training, we combine the NQ and HotpotQA training sets to create a unified dataset for KAGThinker. Training is performed on 8$\times$2 Nvidia H100 GPUs, utilizing full parameter optimization and gradient checkpointing. Key parameter settings are detailed in Table~\ref{tab.training_parameters}.
\renewcommand\arraystretch{1.1}
\begin{table}[htbp]
\centering
\small
\setlength\aboverulesep{0pt}\setlength\belowrulesep{0pt}
\begin{tabular}{p{2.5cm}p{1.0cm}p{2.0cm}p{1.0cm}}
\toprule
\textbf{Parameter}        & \textbf{Value}    & \textbf{Parameter}        & \textbf{Value}   \\ \hline
Learning Rate             & 5e-06    & Weight Decay  & 0.0    \\
Training Batch Size       & 64       & Warmup Ratio  & 0.06   \\
Training Epochs           & 5        & Gradient Clip & 1.0    \\
Max Seq Length            & 16384    & Seed          & 1024   \\
Optimizer                 & Adamw    & Threshold $\tau$& 0.95 \\ \bottomrule
\end{tabular}
\caption{Some important parameters for model training.}
\label{tab.training_parameters}
\end{table}

As presented in Table~\ref{tab.training_samples}, our model utilizes a full training set of a moderate size compared to the baseline. Notably, it demonstrates remarkable data efficiency: even when the training data is scaled down to 1\% (merely a few hundred samples), its performance remains comparable to the current state-of-the-art (cf. Table~\ref{tab.data_ratio}, Sec~\ref{sec.sensitivity_analysis}).
\renewcommand\arraystretch{1.1}
\begin{table}[htbp]
\centering
\small
\setlength\aboverulesep{0pt}\setlength\belowrulesep{0pt}
\begin{tabular}{l|c}
\toprule
              & \textbf{Training Samples} \\ \hline
Naive RAG     & 169,615          \\
R1-Gen        & 169,615          \\
Search-R1     & 169,615          \\
ZeroSearch    & 169,615          \\
StepSearch    & 19,938           \\
ReSearch      & 19,938           \\
Thinker(ours) & 71,116           \\ \bottomrule
\end{tabular}
\caption{Number of training samples for different models.}
\label{tab.training_samples}
\end{table}

We conduct reinforcement learning on the optimal Thinker model using the GRPO algorithm within the VeRL framework. For training, we use the same NQ/HotpotQA dataset as in the SFT stage. The hyperparameters are set with a learning rate of 1e-6, a batch size of 32, a rollout of 5, a maximum search count of 9, a temperature of 1.0, a KL coefficient of 0.001, and a clip ratio of 0.2. The reward function is defined as: (1) a score of 0 for incorrect format; (2) a score of 0.1 for correct format with an F1 score of 0; and (3) the F1 score itself as the reward for a correct answer.

\section{Logical Coherence and Rigor Metrics}
\label{append.logical_metrics}
To evaluate logical coherence and rigor, we have defined four evaluation metrics.

\noindent\textbf{Logical Hierarchy}: Is the decomposition of sub-problems logically structured and easy to follow?

\noindent\textbf{Interleaved Solving}: Are the sub-problems solved in a sequential order, or is there out-of-order execution (e.g., jumping between steps)?

\noindent\textbf{Granularity Consistency}: Is the granularity (i.e., complexity or scope) of sub-problems consistent, or are simple and complex tasks being mixed and executed interchangeably?

\noindent\textbf{Search Efficiency}: Can the sub-problems be directly answered by the the retrieved references?

\section{Analysis of Retrieved Documents}
According to Table~\ref{tab.recall_topk}, the performance gain from retrieving more than two documents per search is negligible. This is because the depth solving module can offset an unsuccessful initial retrieval by conducting subsequent searches.
\renewcommand\arraystretch{1.1}
\begin{table}[htbp]
\small
\centering
\setlength\aboverulesep{0pt}\setlength\belowrulesep{0pt}\setlength{\tabcolsep}{7.5pt}
\begin{tabular}{l|cccccc}
\toprule
          & \textbf{K=1}  & \textbf{K=2}  & \textbf{K=3}  & \textbf{K=4}  & \textbf{K=5}   \\ \midrule
NQ        & 0.395 & 0.417 & 0.450 & 0.444 & 0.452 \\
TriviaQA  & 0.602 & 0.622 & 0.642 & 0.631 & 0.646 \\
PopQA     & 0.444 & 0.463 & 0.484 & 0.473 & 0.483 \\
HotpotQA  & 0.377 & 0.390 & 0.421 & 0.412 & 0.419 \\
2Wiki     & 0.410 & 0.443 & 0.469 & 0.471 & 0.466 \\
MuSiQue   & 0.192 & 0.213 & 0.221 & 0.222 & 0.225 \\
Bamboogle & 0.416 & 0.440 & 0.480 & 0.448 & 0.440 \\ \hline
Avg       & 0.405 & 0.427 & 0.452 & 0.443 & 0.447 \\ \bottomrule
\end{tabular}
\caption{Impact of the number of documents retrieved.}
\label{tab.recall_topk}
\end{table}

\section{Thinker with KAG}
\label{appendix.thinker_with_kag}
\begin{figure}[htbp]
\centering 
\includegraphics[width=0.45
\textwidth]{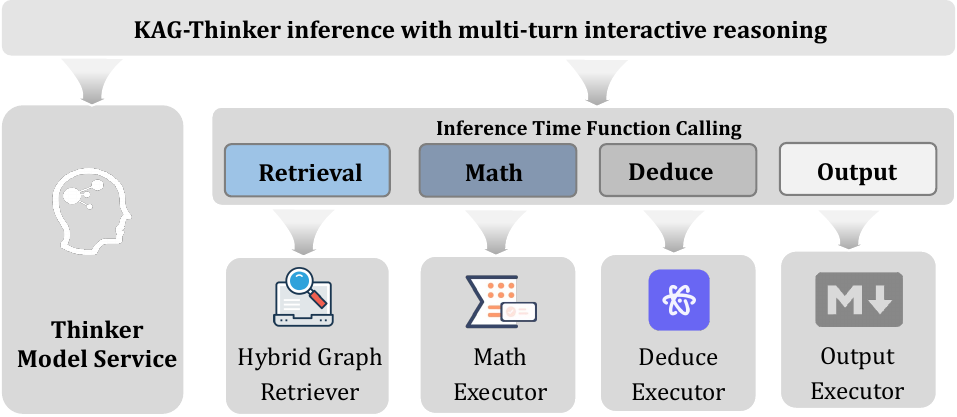} 
\caption{Overview of KAG-Thinker multi-turn interactive thinking and deep reasoning.} 
\label{Fig.thinker_with_kag} 
\end{figure}

The multi-turn interactive reasoning framework of KAG-Thinker is depicted in Figure~\ref{Fig.thinker_with_kag}. We execute the logical form generated by Thinker using the KAG framework's executor, and then feed the resulting output back to Thinker for subsequent tasks. By integrating Thinker with KAG, we can leverage the KAG framework to enhance the utilization of high-quality knowledge bases.

\renewcommand\arraystretch{1.1}
\begin{table*}[ht!]
\centering
\small
\setlength\aboverulesep{0pt}\setlength\belowrulesep{0pt}
\begin{tabular}{c|l|ccc|cccc|c}
\toprule
\multirow{2}{*}{\textbf{LLMs}}                 & \multicolumn{1}{c|}{\multirow{2}{*}{\textbf{Methods}}} & \multicolumn{3}{c|}{\textbf{Single-Hop QA}}                & \multicolumn{4}{c|}{\textbf{Multi-Hop QA}}                                  & \multirow{2}{*}{\textbf{Avg}} \\ \cline{3-9}
                                      & \multicolumn{1}{c|}{}                         &  NQ$^\dagger$      & TriviaQA$^\ast$   & PopQA$^\ast$   & HotpotQA$^\dagger$ & 2Wiki$^\ast$ & MuSiQue$^\ast$ & Bamboogle$^\ast$ &                      \\ \midrule
\multirow{8}{*}{\rotatebox{90}{Qwen2.5-14B-Instruct}} & Naive Generation                             & 0.198          & 0.531          & 0.184          & 0.217          & 0.253          & 0.045          & 0.160          & 0.227                \\
                                      & CoT                                          & 0.190          & 0.495          & 0.148          & 0.269          & 0.297          & 0.054          & 0.432          & 0.269                \\ \cline{2-10}
                                      & Search-o1                                    & 0.347          & 0.635          & 0.241          & 0.268          & 0.161          & 0.099          & 0.416          & 0.310                \\
                                      & IRCoT                                        & 0.114          & 0.375          & 0.166          & 0.230          & 0.248          & 0.102          & 0.312          & 0.221                \\
                                      & Naive RAG                                    & 0.327          & 0.585          & 0.376          & 0.279          & 0.160          & 0.051          & 0.192          & 0.281                \\ \cline{2-10}
                                      & R1-Gen                                       & 0.334          & 0.628          & 0.253          & 0.294          & 0.325          & 0.108          & 0.432          & 0.339                \\
                                      & Search-R1                                    & 0.424          & \textbf{0.660} & 0.442          & \textbf{0.436} & 0.379          & 0.210          & 0.480          & 0.433                \\ \cline{2-10}
                                      & Thinker(ours)                                & \textbf{0.473} & 0.650          & \textbf{0.486} & 0.429          & \textbf{0.487} & \textbf{0.251} & \textbf{0.520} & \textbf{0.471}      \\ \bottomrule
\end{tabular}
\caption{EM performance of different models on Qwen2.5-14B-Instruct. The best performance is set in bold. $^\dagger$/$^\ast$ represents in-domain / out-domain datasets. In contrast to other baselines, StepSearch and ReSearch employ the Musique dataset for training.}
\label{tab.llm14b_results}
\end{table*}

\renewcommand\arraystretch{1.1}
\begin{table*}[ht!]
\centering
\small
\setlength\aboverulesep{0pt}\setlength\belowrulesep{0pt}
\begin{tabular}{l|ccc|cccc|c}
\toprule
\multirow{2}{*}{\textbf{LLMs}}   & \multicolumn{3}{c|}{\textbf{Single-Hop QA}} & \multicolumn{4}{c|}{\textbf{Multi-Hop QA}}       & \multirow{2}{*}{\textbf{Avg}} \\ \cline{2-8}
                        & NQ$^\dagger$      & TriviaQA$^\ast$   & PopQA$^\ast$   & HotpotQA$^\dagger$ & 2Wiki$^\ast$ & MuSiQue$^\ast$ & Bamboogle$^\ast$  &                      \\ \midrule
Qwen3-8B                & 0.445    & 0.626       & 0.464    & 0.403    & 0.450 & 0.227   & 0.448     & 0.438                \\
Llama-3.1-8B-Instruct   & 0.418    & 0.624       & 0.461    & 0.383    & 0.441 & 0.198   & 0.440     & 0.424                \\
Granite-3.3-8B-Instruct & 0.436    & 0.611       & 0.476    & 0.402    & 0.483 & 0.228   & 0.408     & 0.435                \\
Qwen2.5-7B-Instruct     & 0.450    & 0.642       & 0.484    & 0.421    & 0.469 & 0.221   & 0.480     & 0.452   \\ \bottomrule
\end{tabular}
\caption{Performance of Thinker on different LLMs.}
\label{tab.different_llms}
\end{table*}

\renewcommand\arraystretch{1.1}
\begin{table*}[ht!]
\centering
\small
\setlength\aboverulesep{0pt}\setlength\belowrulesep{0pt}
\begin{tabular}{l|ccc|cccc|c}
\toprule
\multirow{2}{*}{\textbf{LLMs}}   & \multicolumn{3}{c|}{\textbf{Single-Hop QA}} & \multicolumn{4}{c|}{\textbf{Multi-Hop QA}}       & \multirow{2}{*}{\textbf{Avg}} \\ \cline{2-8}
                        & NQ$^\dagger$      & TriviaQA$^\ast$   & PopQA$^\ast$   & HotpotQA$^\dagger$ & 2Wiki$^\ast$ & MuSiQue$^\ast$ & Bamboogle$^\ast$  &                      \\ \midrule
Thinker     & 0.450    & 0.642       & 0.484    & 0.421    & 0.469 & 0.221   & 0.480     & 0.452   \\
Thinker-RL  & 0.523    & 0.626       & 0.554    & 0.497    & 0.463 & 0.230   & 0.456     & 0.479   \\ \bottomrule
\end{tabular}
\caption{EM performance of Thinker with RL.}
\label{tab.rl_results}
\end{table*}

\section{Results on 14B LLM}
\label{appendix.14b_llm}

To validate the effectiveness of Thinker on larger-scale base models, we conduct corresponding experiments on Qwen2.5-14B-Instruct. The results, as shown in Table~\ref{tab.llm14b_results}, indicate that our model achieves a 3.7\% improvement in average EM performance across seven datasets compared to Search-R1. It also continues to significantly outperform Search-o1, IRCoT, Naive RAG, and R1-Gen. This demonstrates that our method remains effective on larger-scale models. Meanwhile, compared to the 7B Thinker model, the average performance improved by 1.9\%, with consistent improvements observed across all seven datasets.

\section{Performance of Different LLMs}
\label{appendix.different_llms}

To evaluate the model's effectiveness across different LLMs, a series of experiments are conducted, with the results presented in Table~\ref{tab.different_llms}. The data reveals that while minor performance variations exist among the LLMs, their average performance remains highly comparable. A significant discrepancy is observed on the Bamboogle dataset, a phenomenon primarily attributed to its limited size of only 125 instances. Furthermore, Qwen3-8B is observed to underperform Qwen2.5-7B-Instruct. Our analysis indicates that the model's underperformance stems from high inference length, which increases the likelihood of generating hallucinations or irrelevant content. Nevertheless, the overall performance of Thinker remains robust across the evaluated LLMs, underscoring the strong generalization capability of our constructed dataset.

\section{Performance of Thinker with RL}
\label{appendix.thinker_rl}
We apply the GRPO algorithm to the optimal Thinker model, with the results presented in Table~\ref{tab.rl_results}. As shown, the integration of RL leads to significant performance improvements for Thinker on the NQ, PopQA, and HotpotQA datasets. Our analysis reveals that this is primarily because, during the depth-solving process, RL helps generate more precise search queries and enables a more thorough analysis of the retrieved references.

\section{Training Sample}
\label{appendix.training_sample}
\begin{figure*}[htbp]
\centering 
\includegraphics[width=1.0
\textwidth]{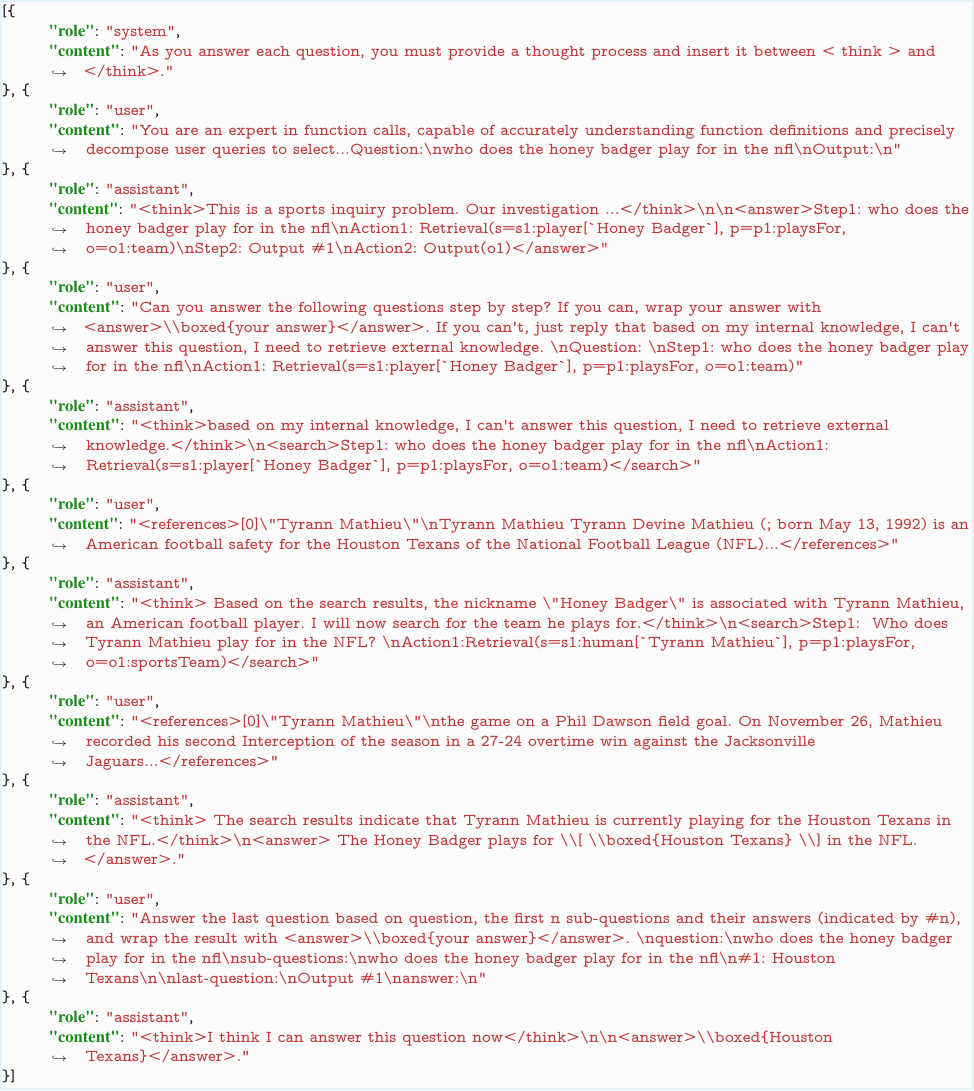} 
\caption{A Thinker training sample supervises the problem-solving process through multi-turn interactions.} 
\label{Fig.train_example} 
\end{figure*}

To more intuitively illustrate our training process, we provide a detailed example, as shown in Figure~\ref{Fig.train_example}. Using the data construction framework illustrated in Figure~\ref{Fig.dataset_construction}, we generate a large-scale dataset of samples for deep search based on hierarchical thinking. These samples are structured as multi-turn interactions. Critically, the reference documents are embedded within the user input and are therefore excluded from the loss function calculation. This approach enhances the model's robustness by compelling it to focus on generating the intended search query rather than merely reproducing the references. Compared to reinforcement learning, this supervised fine-tuning method is significantly faster and more resource-efficient.

\section{Breadth Decomposition}
\label{appendix.breadth_decomposition}
Table~\ref{tab:breadth_decomposition} shows the instruction template for problem breadth decomposition. The primary goal of this process is to decompose a complex problem into n independent and solvable sub-problems.

\renewcommand\arraystretch{1.0}
\begin{table*}[ht!]
\centering
\small
\setlength\aboverulesep{0pt}\setlength\belowrulesep{0pt}
\begin{tabular}{p{0.97\textwidth}}
\toprule
\prompt{You are an expert in function calls, capable of accurately understanding function definitions and precisely decompose user queries to select appropriate functions to solve problems. The functions are as follows:} 
\functionfirst{Function Name: Retrieval}
\functionsecond{\quad Description: Search for SPO information. S stands for subject, O stands for object, represented as variable\_name...}
\functionsecond{\quad Function Usage: Retrieval(s=s\_alias:type[`name'], p=p\_alias:edge, o=o\_alias:type[name], p.prop=`value', s.prop=`value', o.prop=`value')}
\functionfirst{Function Name: Math}
\functionsecond{\quad Description: Perform calculations, which include set operations such as numerical calculations or sorting...}
\functionsecond{\quad Function Usage: Math(content=[`known conditions' or `o\_alias/s\_alias'], target=`goal to be solved')$\to$math\_alias }
\functionfirst{Function Name: Deduce}
\functionsecond{\quad Description: Inference refers to the process of inferring search or calculation results to answer questions...}
\functionsecond{\quad Function Usage: Deduce(op=judgement$|$entailment$|$extract$|$choice$|$multiChoice, content=[`known conditions' or `o\_alias/s\_alias'], target=`goal to be solved')$\to$deduce\_alias}
\functionfirst{Function Name: Output}
\functionsecond{\quad Description: Directly output A, B, ... as the answer, where A and B are variable names referencing previous...}
\functionsecond{\quad Function Usage: Output(A,B,...)}
\prompt{Please, based on the definition of the above function, decompose the user question into one or multiple logical steps, outputting the execution plan for each step along with the corresponding action. Please note:
Step: Accurately point out the logical thinking process of the question, and use \#1 to refer to the solution result of Step1, \#2 to refer to the solution result of Step2, and so on
Action: Indicate exactly the function you selected and its parameters.} \hline
\breadthquestion{\textbf{Question:} Which film has the director who died first, Hit Parade Of 1947 or Khiladi 420?} \hline
\breadthanswer{\textbf{Output:} \textcolor{red}{$<$think$>$}This is a comparison problem. To address the question of which film, between ``Hit Parade Of 1947" and ``Khiladi 420" was directed by the director who passed away first, we will undertake a chronological investigation. Our initial step involves identifying the director of ``Hit Parade Of 1947". Following this, we will ascertain the director of ``Khiladi 420". With both directors' identities established, we will then determine which of these individuals died earlier. This chronological analysis will enable us to connect the director's lifespan to the films they directed, ultimately revealing which film was associated with the director who died first. By meticulously comparing the timelines of the directors' lives, we ensure that our conclusion is based on accurate historical data.\textcolor{red}{$<$/think$>$}\texttt{$\backslash$n}\texttt{$\backslash$n}\textcolor{teal}{$<$answer$>$}Step1: Who is the director of Hit Parade Of 1947?\texttt{$\backslash$n}Action1: Retrieval(s=s1:film[`Hit Parade Of 1947'], p=p1:director, o=o1:director)\texttt{$\backslash$n}Step2: When did \#1 die?\texttt{$\backslash$n}Action2: Retrieval(s=o1, p=p2:deathtime, o=o2:deathtime)\texttt{$\backslash$n}Step3: Who is the director of Khiladi 420?\texttt{$\backslash$n}Action3: Retrieval(s=s3:film[`Khiladi 420'], p=p3:director, o=o3:director)\texttt{$\backslash$n}Step4: When did \#3 die?\texttt{$\backslash$n}Action4: Retrieval(s=o3, p=p4:deathtime, o=o4:deathtime)\texttt{$\backslash$n}Step5: Which film was directed by the director who died first according to \#2 and \#4?\texttt{$\backslash$n}Action5: Deduce(op=choice, content=[`o2',`o4'], target=`which film was directed by the director who died first according to o2 and o4')$\to$o5 \textcolor{teal}{$<$/answer$>$}}
\bottomrule
\end{tabular}
\caption{Instruction template for breadth decomposition. It initiates with problem positioning and analysis to determine decomposition steps, ultimately generating all atomic sub-problems through single-pass decomposition. Sub-question dependencies are propagated via $\#n$ in natural language Steps and referenced through logical function variables (e.g., $o_n$, $s_n$) in Actions.}
\label{tab:breadth_decomposition}
\end{table*}

\section{Knowledge Boundary Determination}
\label{appendix.knowledge_boundary_determination}
To decide whether a deep search is required, we first test the LLM's knowledge boundary by prompting it to answer the sub-problem directly (template in Table~\ref{table:prompt_answer}). Subsequently, the LLM evaluates its own response to determine if it is adequate, as guided by the instruction template in Table~\ref{table:prompt_assess}.

\renewcommand\arraystretch{1.1}
\begin{table}[htbp]
    \centering 
    \small
    \begin{tabular}{p{8cm}}
        \hline
        \prompt{Can you answer the following question step by step?}
        \prompt{If you can, wrap your answer with $<$answer$>$\textbackslash\textbackslash boxed\{your answer\}$<$/answer$>$.}
        \prompt{If you can't, just reply that based on my internal knowledge, I can't answer this question, I need to retrieve external knowledge.}
        \breadthquestion{\textbf{Question}: Who is the author of The Pequod?}
        \hline
    \end{tabular}
    \caption{Instruction template for solving sub-problem.}\label{table:prompt_answer}
\end{table}

\renewcommand\arraystretch{1.1}
\begin{table}[htbp]
    \centering 
    \small
    \begin{tabular}{p{8cm}}
        \hline
        \prompt{Analyze whether the answer to the following question is correct?}
        \prompt{Please think step-by-step before arriving at your conclusion.}
        \prompt{If yes, answer True; if no, answer False.}
        \prompt{Wrap your answer with $<$answer$>$\textbackslash\textbackslash boxed\{answer\}$<$/answer$>$.}
        \breadthquestion{\textbf{Question:} Who is the author of The Pequod?}
        \breadthanswer{\textbf{Answer:} Herman Melville}
        \hline
    \end{tabular}
    \caption{Instruction template for confidence assessment. Where Question represents the sub-problem to be solved, and Answer represents the LLM-generated answer.}\label{table:prompt_assess}
\end{table}

\section{Focusing and Reasoning}
\label{appendix.focusing_and_reasoning}

Table~\ref{tab:focusing_and_reasoning} shows an example of the data construction for focusing and reasoning. In this task, the LLM is instructed to analyze the current question and the retrieved references. It must then provide a judgment on whether these references are sufficient to answer the question, along with a justification for its assessment.

\begin{table*}[ht!]
\centering
\small
\setlength\aboverulesep{0pt}\setlength\belowrulesep{0pt}
\begin{tabular}{p{0.97\textwidth}}
\toprule
\prompt{According to the question and references, analyze the relationship between each reference and the question, then summarize whether these references can answer the above questions and give reasons. Wrap the answer and the corresponding reasons in \textcolor{teal}{$<$answer$>$}Yes/No\textcolor{teal}{$<$/answer$>$} and \textcolor{orange}{$<$reason$>$}your reason\textcolor{orange}{$<$/reason$>$}. }\hline
\depthreference{\textbf{References:}}
\depthreference{0. ``Splash (U.S. TV series)" Splash (U.S. TV series) Splash is an American competition-based reality show with celebrity diving competitions which broadcast on ABC from March 19 to May 7, 2013. It was based on the Dutch reality franchise ``Celebrity Splash!" created by Eyeworks for the series ``Stars Jumping on Saturday" that premiered in 2012. The show was hosted by actor Joey Lawrence and sportscaster Charissa Thompson with former Olympic divers Steve Foley from Australia and David Boudia from the U.S. as judges...}
\depthreference{1. ``Celebrity Splash!" Celebrity Splash! Celebrity Splash! is a reality television franchise created by Dutch company Eyeworks, started from their Dutch reality show ``Sterren Springen Op Zaterdag" which premièred in 2012. The franchise involves celebrities diving into the pool. ``Splash" has its origin and idea from German Olympic-themed variety TV show `` (``TV total Diving"), it was first aired on 16 December 2004, in the ``TV total" show, on ProSieben and was founded by Stefan Raab and hosted by Sonya Kraus...}
\depthreference{2. ``Celebrity Splash! (Argentinian TV series)" Celebrity Splash! (Argentinian TV series) Splash! is a reality television show which teaches celebrities the art of diving. The first series is featured to start on 11 June 2013, and it will be broadcast by Telefe. Marley Wiebe will be hosting the show, with Pampita Ardohain, Maximiliano Guerra, Miguel Ángel Rodríguez and Mariana Montes as judges. ``Splash" has its origin and idea from German TV Olympic-themed variety show (``TV Total Diving"), it was first aired on 16 December 2004, in the TV total slot, on ProSieben and was founded by Stefan Raab and hosted by Sonya Kraus. Other hosts/reporters...} \hline
\breadthquestion{\textbf{Question:} Who is the host of Splash!} \hline
\breadthanswer{\textbf{Output:}} 
\breadthanswer{\textcolor{teal}{$<$answer$>$}Yes\textcolor{teal}{$<$/answer$>$}}
\breadthanswer{\textcolor{orange}{$<$reason$>$}The host of the U.S. TV series Splash is actor Joey Lawrence, alongside sportscaster Charissa Thompson. This is explicitly confirmed in reference 0, which details the U.S. version of the show. References 1 and 2 discuss the Dutch and Argentinian iterations of the franchise, which have different hosts (e.g., Sonya Kraus, Marley Wiebe) and are unrelated to the U.S. version.\textcolor{orange}{$<$/reason$>$}} 
\bottomrule
\end{tabular}
\caption{Focusing and Reasoning data construction example: By analyzing and evaluating whether the references can sufficiently address the sub-question, the system determines whether to proceed to the next Action or directly generate the final answer.}
\label{tab:focusing_and_reasoning}
\end{table*}

\section{Medical Field Application}
\label{appendix.medical_application}
\noindent\textbf{Med-Thinker Construction}. To evaluate the generalization ability of the Thinker model, we develop Med-Thinker for the medical domain based on the Thinker framework. We use a breadth decomposition approach to decompose the problem into multiple sub-problems, and we also employ knowledge boundary determination module to determine whether retrieval is necessary. Table \ref{tab:med_planner} illustrates the reasoning process of Med-Thinker. First, within the breadth decomposition part, the content between {$<$think$>$} and {$<$/think$>$} represents the thought process. Here, it is reasoned that calculating \textit{the required fluid volume} involves considering both \textit{the burn area and weight}. Consequently, information regarding \textit{the percentage of lower limb (including buttock) surface area in children}, as well as \textit{the ratio of colloidal fluids to crystalloid fluids}, should be retrieved for further decomposition of the sub-problems. Second, in the depth solving part, each sub-problem is addressed independently. For sub-problems related to retrieval, the system first performs its own reasoning and then evaluates whether the answer is reliable and complete. If not, it generates a retrieval directive using $<$search$>$...$<$/search$>$. The Chunks retrieved for each sub-problem (with top-k=3) are integrated into the context. Since the retrieved Chunks may be lengthy, we perform knowledge compression and information targeting on the results for each sub-problem. The answers are then refined on the basis of the reference. For sub-problems requiring retrieval (e.g., Step 4 and Step 5), a deep retrieval check is employed to assess whether the Reference can resolve the sub-problem. If the Reference cannot address the sub-problem, a new retrieval directive is generated. Specifically, for Step 6, the Math directive is executed by generating Python code and obtaining computational results using a code interpreter. Finally, based on the breadth decomposition and depth solving, the complete reasoning process and the final result are generated.

\noindent\textbf{Dataset}. We use MedQA as our evaluation dataset. This dataset comprises question-and-answer pairs from professional medical examinations. All questions are multiple-choice, covering topics such as medical knowledge, disease diagnosis, pharmacology, and diagnostic procedures. As the original dataset is extensive, we randomly sample 300 entries from its Chinese subset for our evaluation.

\noindent\textbf{Implementation Details}. We perform SFT on three base models: Llama3.1-8B-Instruct, Qwen2.5-14B-Instruct and DeepSeek-R1-Distill-Qwen-14B. The training is configured with a learning rate of 5e-6 and 5 epochs. For this process, we utilize a self-constructed medical knowledge base, for which vector indexes are built using BGE-M3.

\renewcommand\arraystretch{1.1}
\begin{table}[htbp]
\centering
\small
\setlength\aboverulesep{0pt}\setlength\belowrulesep{0pt}
\begin{tabular}{p{1cm}<{\centering}|p{4.0cm}|p{2.0cm}<{\centering}}
\toprule
\textbf{LLMs} & \textbf{Methods} & \textbf{MedQA}   \\ \midrule
\multirow{6}{*}{\rotatebox{90}{Llama3.1-8B}}  & Naive Generation & 58.33  \\ 
    & Naive RAG & 61.54  \\ 
    & Naive RAG+adaptive   & 61.33   \\ 
    & IRCoT   & 62.67   \\ 
    & ReAct   & 39.22   \\ 
    & Med-Thinker(ours)   & \textbf{74.00}  \\
\hline
\multirow{6}{*}{\rotatebox{90}{Qwen2.5-14B}}  & Naive Generation & 84.00  \\ 
    & Naive RAG & 81.67  \\ 
    & Naive RAG+adaptive   & 85.00   \\ 
    & IRCoT   & 82.33   \\ 
    & ReAct   & 86.20   \\ 
    & Med-Thinker(ours)    & \textbf{87.00}  \\ \hline
\multirow{6}{*}{\rotatebox{90}{DeepSeek-R1-14B}}  & Naive Generation & 79.67  \\ 
    & Naive RAG & 79.00  \\ 
    & Naive RAG+adaptive   & 81.48  \\ 
    & IRCoT   & 81.33   \\ 
    & ReAct   & 80.87   \\ 
    & Med-Thinker(ours)   & \textbf{85.28}  \\
\bottomrule
\end{tabular}
\caption{Accuracy of different models on the MedQA dataset. The best performance is set in bold.}
\label{tab:med-result}
\end{table}

\renewcommand\arraystretch{1.0}
\begin{table*}[ht!]
\centering
\small
\setlength\aboverulesep{0pt}\setlength\belowrulesep{0pt}
\begin{tabular}{p{0.97\textwidth}}
\toprule
\breadthquestion{\textbf{Question}: Boy, 3 years old, weighing 16kg, sustained scald burns on both lower limbs from hot water. Physical examination: BP 85/60 mmHg, restless. Both lower limbs (including buttocks) suffered second-degree burns, with a urine output of 15ml/h. The amount of colloid fluid that should be administered in the first 24 hours is (    ). Options: A: 500ml B: 800ml C: 400ml D: 600ml E: 700ml} 
\hline
\functionfirst{\textbf{Breadth Decomposition}: 

\textbf{$<$think$>$} The user's question concerns a 3-year-old boy weighing 16kg...

First, we need to calculate the fluid volume based on the burn area and the child's weight...The proportion of colloid fluid to crystalloid fluid is typically 1:1 or 0.5:1, depending on the severity of the burn. Next, we need to clarify several key aspects: ... \textbf{$<$/think$>$}

\textbf{Step 1: Fluid Requirement Formula}

Action 1: Retrieval(s=s1:Medical formula[`Fluid Requirement Formula'], p=p1, o=o1)

\textbf{Step 2: Fluid Replacement Formula for the First 24 Hours}

Action 2: Retrieval(s=s1, p=p2:time, o=o2:time[`The first 24 hours'])

\textbf{Step 3: The Percentage of Total Body Surface Area (TBSA) for Both Lower Limbs (Including the Buttocks)}

Action 3: Retrieval(s=s2:BodyParts[`Both Lower Limbs (Including the Buttocks)'], p=p3:AreaRatio, o=o3:AreaRatio)

\textbf{Step 4: The Percentage of Total Body Surface Area (TBSA) for Both Lower Limbs (Including the Buttocks) in Children}

Action 4: Retrieval(s=s2, p=p4:BelongTo, o=o4:Crowd[`Children'])

\textbf{Step 5: The Proportion of Colloid Fluid in the Total Fluid Volume for Second-Degree Burns}

Action 5: Retrieval(s=s3:TreatmentPlan[`Burn fluid replacement'], p=p5:Colloidal liquid ratio, o=o5:Colloidal liquid ratio, p.Degree of burn=II$^{\circ}$)

\textbf{Step 6: Based on \#1, \#2, \#3, \#4, and \#5, calculate the amount of colloid fluid to be administered in the first 24 hours.}

Action 6: Math(content=[`o1', `o2', `o3', `o4', `o5', `weight=16kg'], target=`The amount of colloid fluid to be administered in the first 24 hours')$\to$math6
} \hline

\functionfirst{\textbf{Depth Solving}:

\textbf{Subanswer 1}: The formula for calculating fluid replacement in burn patients is often based on the Parkland formula...

\textbf{Subanswer 2}: According to burn formulas, the basic principle for fluid resuscitation in pediatric burns is:

Total fluid volume for the first 24 hours = Body weight (kg) $\times$ Burned percentage (\%TBSA) $\times$ 1.5 (pediatric coefficient) + baseline physiological fluid requirement.

\textbf{Subanswer 3}: The burn area of both lower limbs (including the buttocks) accounts for approximately 46\% of the Total Body Surface Area (TBSA) in adults...

\textbf{Subanswer 4}: The burned area for both lower limbs (including buttocks) accounts for approximately \textcolor{red}{46\%} of TBSA in adults, but the percentage may differ for children and needs to be further verified. \textcolor{blue}{$<$search$>$Step: The Percentage of Total Body Surface Area (TBSA) for Both Lower Limbs (Including the Buttocks) in Children. Action: Retrieval(s=s2, p=p4:BelongTo, o=o4:Crowd[`Children'])$<$/search$>$}

$<$Reference$>$...$<$/Reference$>$

\textcolor{teal}{The area of both lower limbs (including the buttocks) for children is calculated as 46\% - (12 - age)\% = 37\%.}

\textbf{Subanswer 5}: Parkland Formula: \textcolor{red}{1ml} of fluid per kilogram of body weight per 1\% of burned area. Half of the total fluid volume is colloid fluid. However, considering the requirements for the first 24 hours and the differences in pediatric cases, I need to confirm further. \textcolor{blue}{$<$search$>$Step: The Proportion of Colloid Fluid in the Total Fluid Volume for Second-Degree Burns. Action: Retrieval(s=s3:TreatmentPlan[`Burn fluid replacement'], p=p5:Colloidal liquid ratio, o=o5:Colloidal liquid ratio, p.Degree of burn=II$^{\circ}$)$<$/search$>$}

$<$Reference$>$...$<$/Reference$>$

\textcolor{teal}{According to the 7th Edition of the Surgery Fluid Resuscitation Protocol, the amount of colloid and electrolyte solution to be provided for pediatric patients is 2.0ml, with a ratio of 1:1 between colloid and electrolyte solution.}

\textbf{Subanswer 6}: $<$Python$>$...$<$/Python$>$

\textcolor{teal}{592ml}
} \hline
\breadthanswer{\textbf{Answer}:

1.Determine Burn Area:

The patient is ..., the total burn area is calculated to be 37\%.

2.Calculate Colloid Volume:

According to the fluid replacement protocol ... Colloid volume = Body weight (kg) × Total burn area (\%) × 1
= 16 kg × 37\% × 1 = 592 ml.

Since 592 ml is approximately 600 ml, option D is selected.
}
\bottomrule
\end{tabular}
\caption{An example of the problem-solving process in Med-Thinker. The breadth decomposition module first generates the thought process, then decomposes the problem into multiple sub-problems. In the depth solving module, an adaptive retrieval equilibrium model is used for reasoning and external knowledge retrieval. Finally, the answer is generated based on the context. The red font part represents errors in the questions generated by the model, the blue font part represents adaptive generation of retrieval instructions, and the green font part represents the results executed by logical form executors.}
\label{tab:med_planner}
\end{table*}

\renewcommand\arraystretch{1.1}
\begin{table*}[htbp]
\centering
\small
\setlength\aboverulesep{0pt}\setlength\belowrulesep{0pt}
\begin{tabular}{lccccc}
\toprule
          & Samples*Epochs & Training Cost (h) & Avg EM score & Avg inference cost (s/sample) & Avg \#retrievals/sample \\ \midrule
Search-R1 & 169615*2       & 90                & 0.385        & 0.28                          & 3.60                    \\ \hline
ReSearch  & 19938*2        & 71                & 0.411        & 0.17                          & 3.73                    \\ \hline
Thinker   & 71116*5        & 24                & 0.452        & 0.34                          & 3.58  \\  \bottomrule                
\end{tabular}
\caption{Computational cost of training and inference for different models.}
\label{tab.computational_cost}
\end{table*}
\noindent\textbf{Experimental Results}. The empirical evaluation results are presented in Table \ref{tab:med-result}. Med-Thinker consistently demonstrates enhanced performance across both base models, with a particularly significant improvement being observed on the Llama3.1-8B-Instruct. In particular, compared to the established multi-turn planning and retrieval augmented models, IRCoT and ReAct, Med-Thinker achieves remarkable performance improvements of 11.33\%, 34.78\% respectively. It also outperforms the Naive RAG, by a margin of 12.46\%. Among them, ReAct struggles to follow the instruction format on Llama3.1-8B-Instruct, resulting in poorer performance. On the DeepSeek-R1-Distill-Qwen-14B, Med-Thinker shows similar superiority, outperforming IRCoT, ReAct, and adaptive Naive RAG model by 3.95\%, 4.41\%, and 3.8\%, respectively. The baselines based on Qwen2.5-14B-Instruct have already achieved good performance, but Med-Thinker still outperforms other reasoning and RAG models, achieving the best results. These consistent gains across different base models and comparative baselines unequivocally validate the effectiveness and robustness of Med-Thinker.
\section{Computational Cost}
We use the same GPUs for Search-R1, ReSearch, and Thinker: 8×A100-80G. Following the official open-source configurations, we adjust the batch size during training to maximize resource utilization; other parameters remain consistent with those described in their paper. During inference, we employ multi-threaded inference with 200 threads to ensure maximum resource utilization. The computational costs for different models are shown in Table~\ref{tab.computational_cost}.

From the table, we see that Thinker's training cost is significantly lower than the baseline methods. During inference, Thinker's higher cost is attributed to its architecture, which introduces a dual representation and a logical function, resulting in higher token consumption. By incorporating knowledge boundary determination, the model’s average number of retrievals does not increase and is comparable to the other methods.
\end{document}